\documentclass[accepted]{uai2022} 

\usepackage[american]{babel}

\usepackage{natbib} 
\usepackage{mathtools} 
\usepackage{booktabs} 
\usepackage{tikz} 
\usepackage{multirow}
\usepackage{diagbox}
\usepackage{bm}
\usepackage{graphicx}
\usepackage{amsmath}
\usepackage{amssymb}
\usepackage{times}
\usepackage{epsfig}
\usepackage{microtype} 
\usepackage{makecell}
\newtheorem{definition}{Definition}

\title{Encouraging Disentangled and Convex Representation with Controllable
Interpolation Regularization}

\author{Yunhao Ge, Zhi Xu, Yao Xiao, Gan Xin, Yunkui Pang, Laurent Itti \\
University of Southern California, Los Angeles, CA, USA\\
\texttt{yunhaoge@usc.edu, itti@usc.edu}
}
  
\begin{document}
\maketitle
\vspace{-25pt}
\begin{abstract}
\vspace{-20pt}
We focus on controllable disentangled representation learning (C-Dis-RL), where users can control the partition of the disentangled latent space to factorize dataset attributes (concepts) for downstream tasks. Two general problems remain under-explored in current methods: (1) They lack comprehensive disentanglement constraints, especially missing the minimization of mutual information between different attributes across latent and observation domains. (2) They lack convexity constraints, which is important for meaningfully manipulating specific attributes for downstream tasks. To encourage both comprehensive C-Dis-RL and convexity simultaneously, we propose a simple yet efficient method: Controllable Interpolation Regularization (CIR), which creates a positive loop where disentanglement and convexity can help each other. Specifically, we conduct controlled interpolation in latent space during training, and we reuse the encoder to help form a 'perfect disentanglement' regularization. In that case, (a) disentanglement loss implicitly enlarges the potential understandable distribution to encourage convexity; (b) convexity can in turn improve robust and precise disentanglement. 
CIR is a general module and we merge CIR with three different algorithms: ELEGANT, I2I-Dis, and GZS-Net to show the compatibility and effectiveness. Qualitative and quantitative experiments show improvement in C-Dis-RL and latent convexity by CIR. This further improves downstream tasks: controllable image synthesis, cross-modality image translation and zero-shot synthesis. 
\end{abstract}
\vspace{-10pt} 
\section{Introduction}
\vspace{-6pt}





\begin{figure}[t]
\vspace{-5pt}
\begin{center}
\includegraphics[width=\linewidth]{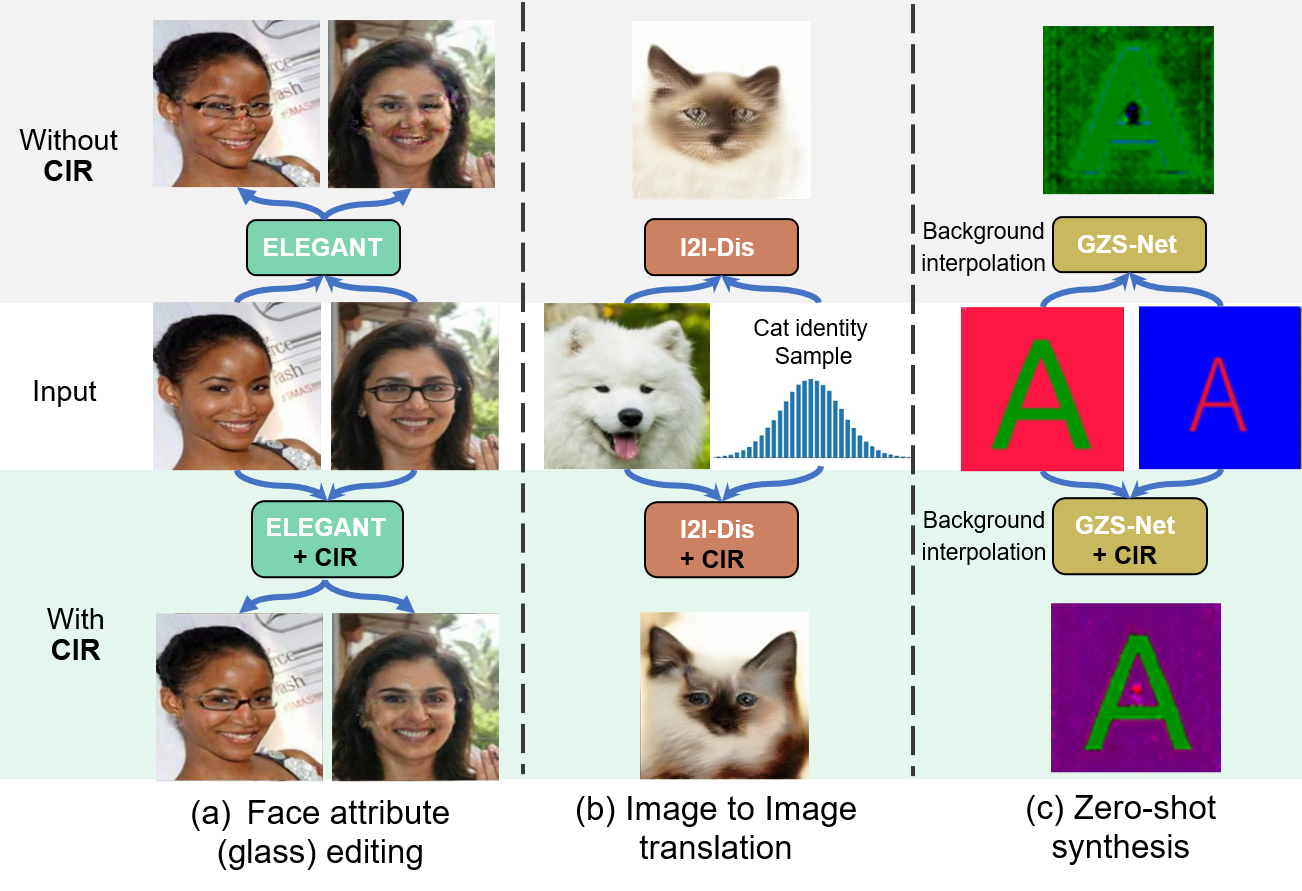}
\end{center}
\setlength{\abovecaptionskip}{-1pt}
  \caption{Our proposed approach CIR improves the result quality of 3 tasks by encouraging both disentanglement and convexity in the latent space: (a) Face attribute editing with ELEGANT (add/remove glasses on face); CIR is better able to transfer glasses with less disturbance on other face parts. (b) Image to image translation transfer from a dog image to a cat image with same pose (content); CIR better matches the desired pose with fewer artifacts. (c) Zero-shot synthesis with GZS-Net to synthesize an image with a new background by interpolating in the corresponding latent space; CIR better interpolates the background only without changing letter size, color or font style. See Suppl.\ fig.\ 1 for a larger version.}
  

\label{fig:1}
\vspace{-10pt} 
\end{figure}


Disentangled representation learning empowers models to learn an orderly latent representation, in which each separate set of dimensions is responsible for one semantic attribute \cite{higgins2016beta, chen2016infogan, zheng2019joint}. 
If we categorize different disentangled representation methods by whether they could  \textit{control} the partition of the obtained disentangled latent representation (e.g., explicitly assign first 10 dimensions to be responsible for face attribute), there are two main threads: 

(1) \textbf{Uncontrollable} disentangled methods, such as Variational Autoencoders (VAEs) \cite{kingma2014autoencoding,Higgins2017betaVAELB,tran2017disentangled}, add prior constraints (e.g., Gaussian distribution)  in latent space to implicitly infer a disentangled latent code. Most are unsupervised methods that can easily generalize to different datasets and extract latent semantic factors. Yet, they struggle to obtain controllable disentanglement because the unsupervised latent encoding does not map onto user-controllable attributes. 
(2) \textbf{Controllable} disentangled methods, which explicitly control the partition of the disentangled latent space and the corresponding mapping to semantic attributes by utilizing dataset attribute labels or task domain knowledge. Because users can precisely control and design their task-driven disentangled latent representation, they are widely used in various downstream tasks: in cross-modality image-to-image translation, I2I-Dis \cite{lee2018diverse} disentangle content and attribute to improve image translation quality (Fig.~\ref{fig:1}(b)); In controllable image synthesis, ELEGANT \cite{xiao2018elegant} and DNA-GAN \cite{xiao2017dna} disentangle different face attributes to achieves face attribute transfer by exchanging certain part of their latent encoding across images ((Fig.~\ref{fig:1}(a))). In group supervised learning, GZS-Net \cite{ge2020zero} uses disentangled representation learning to simulate human imagination and achieve zero-shot synthesis (Fig.~\ref{fig:1}(c)).


However, controllable disentangled methods suffer from 2 general problems:
1) The constraints on disentanglement are partial and incomplete, they lack \textit{comprehensive} disentanglement constraints. For example, while ELEGANT enforces that modifying the part of the latent code assigned to an attribute (e.g., hair color) will affect that attribute, it does not explicitly enforce that a given attribute will {\em not} be affected when the latent dimensions for other attributes are changed (Fig.~\ref{fig:1}(a)). 2) Most of the above-mentioned downstream tasks require manipulating specific attribute-related dimensions in the obtained disentangled representation; for instance, changing only the style while preserving the content in an image-to-image translation task. For such manipulation, the convexity of each disentangled attribute representation (i.e., interpolation within that attribute should give rise to meaningful outputs) is not guaranteed by current methods (Fig.~\ref{fig:1}, Fig.~\ref{fig:3}(a) and Fig.~\ref{fig:7}(a)). Further, convexity demonstrates an ability to generalize, which implies that the autoencoder structure has not simply memorized the representation of a small collection of data points. Instead, the model uncovered some structure about the data and has captured it in the latent space \cite{berthelot2018understanding}. 
How to achieve both comprehensive disentanglement, and convexity in the latent space, is under-explored.

To solve the above problems, we first provide a definition of controllable disentanglement with the final goals of \textit{perfect} controllable disentanglement and of convexity in latent space. Then, we use information theory and interpolation to analyze different ways to achieve disentangled (Sec.~\ref{sec:3.1}) and convex (Sec.~\ref{sec:3.2}) representation learning. To optimize them together, based on the definition and analysis, we use approximations to create a positive loop where disentanglement and convexity can help each other. We propose Controllable Interpolation Regularization (CIR), a simple yet effective general method that compatible with different algorithms to encourage both controllable disentanglement and convexity in the latent space (Sec.~\ref{sec:3.3}). Specifically, 
CIR first conducts controllable interpolation, i.e., controls which attribute to interpolate and how in the disentangled latent space, then reuses the encoder to 're-obtain' the latent code and add regularization to explicitly encourage \textit{perfect} controllable disentanglement and implicitly boost convexity. We show that this iterative approximation approach converges towards perfect disentanglement and convexity in the limit of infinite interpolated samples. 

Our contributions are: 
(i) Describe a new abstract framework for \textit{perfect} controllable disentanglement and convexity in the latent space, and use information theory to summarize potential optimization methods (Sec.~\ref{sec:3.1}, Sec.~\ref{sec:3.2}).
(ii) Propose Controllable Interpolation Regularization (CIR), a general module compatible with different algorithms, to encourage both controllable disentanglement and convex in latent representation by creating a positive loop to make them help each other. CIR is shown to converge towards perfect disentanglement and convexity for infinite interpolated samples (Sec.~\ref{sec:3.3}).
(iii) Demonstrate that better disentanglement and convexity are achieved with CIR on various tasks: controllable image synthesis, cross-domain image-to-image translation and group supervised learning (Sec.~\ref{sec:4}, Sec.~\ref{sec:5}). 


\vspace{-12pt}
\section{Related Work}
\vspace{-8pt}
\noindent{\bf {Controllable Disentangled Representation Learning}} (C-Dis-RL)
is different from Uncontrollable Dis-RL (such as VAEs \cite{kingma2014autoencoding,Higgins2017betaVAELB,chen2018isolating}), which implicitly achieves disentanglement by incorporating a distance measure into the objective, encouraging the latent factors to be statistically independent. However, these methods and not able to freely control the relationship between attribute and latent dimensions.
C-Dis-RL learns a partition control of the disentanglement from semantic attribute labels in the latent representation and boosts the performance of various tasks: ELEGANT \cite{xiao2018elegant} and DNA-GAN \cite{xiao2017dna} for face attribute transfer; I2I-Dis \cite{lee2018diverse} for diverse image-to-image translation; DGNet \cite{zheng2019joint} and IS-GAN \cite{eom2019learning} for person re-identification;  GZS-Net \cite{ge2020zero} for controllable zero-shot synthesis with group-supervised learning. However, their constraints on disentanglement are implicit and surrogate by image quality loss, which also misses the constraint between different attributes across latent and observation. 
As a general module, CIR is compatible and complementary with different C-Dis-RL algorithms by directly constraining disentanglement while focusing on minimizing the mutual information between different attributes across latent and observation.\\ 
\noindent{\bf Convexity of Latent Space} is defined as a set in which the line segment connecting any pair of points will fall within the rest of the set \cite{sainburg2018generative}. Linear interpolations in a low-dimensional latent space often produce comprehensible representations when projected back into high-dimensional space \cite{engel2017neural, ge2020pose}.
However, linear interpolations are not necessarily justified in many controllable disentanglement models because latent-space projections are not trained explicitly to form a convex set. VAEs overcome non-convexity by forcing the latent representation into a pre-defined distribution, which may be a suboptimal representation of the high-dimensional dataset. GAIN \cite{sainburg2018generative} adds interpolation in the generator in the middle latent space and uses a discriminative loss on a GAN structure to help optimize convexity. Our method controls the interpolation in a subspace of the disentangled latent space and uses disentanglement regularization to encourage a convex latent space for each semantic attribute.
\vspace{-10pt}
\section{Controllable Interpolation Regularization}
\label{sec:3}
\vspace{-8pt}
\subsection{Mutual Information for Perfect Controllable disentanglement }
\vspace{-5pt}
\label{sec:3.1}

\begin{figure}[t]
\vspace{-15pt}
\begin{center}
\includegraphics[width=\linewidth]{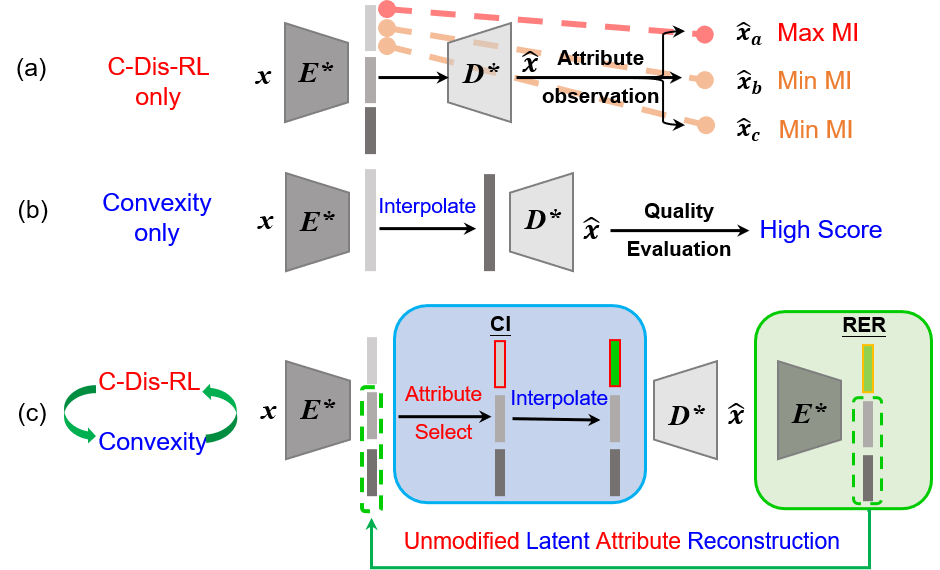}
\end{center}
\setlength{\abovecaptionskip}{-1pt}
  \caption{Intuitive understanding of Controllable Interpolation Regularization (\textbf{CIR}). (a) Only encourage \textcolor{red}{controllable disentangled representation} (C-Dis) with general Mutual Information (MI) constrain method: maximize the MI between the same attribute across latent and observation domains while minimizing the MI between the different attribute across latent and observation domains.(b) Only encourage  \textcolor{blue}{convexity} with interpolation and image quality evaluation. (c) A simple yet efficient method, CIR, encourages both C-Dis and convexity in latent representation.
  CIR consists of a Controllable Interpolation (CI) module and a Reuse Encoder Regularization (RER) module.}
\label{fig:2}
\vspace{-10pt}
\end{figure}

A general autoencoder structure $(D \circ E)$: $\mathcal{X} \rightarrow \mathcal{X}$  is composed of an encoder network $E : \mathcal{X} \rightarrow \mathbb{R}^d$, and a decoder network $D : \mathbb{R}^d  \rightarrow \mathcal{X}$. $\mathbb{R}^d$ is a latent space, compared with the original input space $\mathcal{X}$ (e.g., image space). The disentanglement is a property of latent space $\mathbb{R}^d$ where each separate set of dimensions is responsible for one semantic attribute of given dataset. 
Formally, a dataset (e.g., face dataset) contains $n$ samples $\mathcal{D} = \{x^{(i)}\}_{i=1}^n$, each accompanied by $m$ attributes
$\mathcal{D}_a = \{(a_1^{(i)}, a_2^{(i)}, \dots a_m^{(i)}) \}_{i=1}^n$. Each attribute $a_j \in \mathcal{A}_j$ can be either binary (two attribute values, e.g., $\mathcal{A}_1$ may denote wearing glass or not; $\mathcal{A}_1 = \{\textrm{wear glass}, \textrm{not wear glass}$\}), or a multi-class attribute, which contains a countable set of attribute values (e.g., $\mathcal{A}_2$ may denote hair-colors $\mathcal{A}_2 = \{\textrm{black}, \textrm{gold}, \textrm{red}, \dots \}$). 
Controllable disentangled representation learning (C-Dis-RL) methods have two properties: (1) Users can explicitly control the partition of the disentangled latent space $\mathbb{R}^d$ and (2) Users can control the semantic attributes mapping between $\mathbb{R}^d$ to input space $\mathcal{X}$. To describe the ideal goal for all C-Dis-RL, we define a \textit{perfect} controllable disentanglement property in latent space $\mathbb{R}^d$ and the autoencoder.

\begin{definition}
\textsc{\textit{perfect} controllable disentanglement (\textit{perfect}-C-D)}$(E, D, \mathcal{D})$: Given a general encoder $E : \mathcal{X} \rightarrow \mathbb{R}^d$, a decoder $D : \mathbb{R}^d  \rightarrow \mathcal{X}$, and a dataset $\mathcal{D}$ with $m$ independent semantic attributes $\mathcal{A}$,
we say the general autoencoder achieve \textit{perfect} controllable disentanglement for dataset $\mathcal{D}$ if the following property is satisfied:
(1) For encoder $E$, if one attribute $\mathcal{A}_i$ of input $x$ was specifically modified, transforming $x$ into $\hat{x}$, after computing latent codes $z = E(x)$ and $\hat{z} = E(\hat{x})$, the difference between $z$ and $\hat{z}$ should be zero for all latent dimensions except those that represent the modified attribute. (2) Similarly, for decoder $D$, the latent space change should only influence the corresponding attribute expression in the output (e.g., image) space. 
\end{definition}

To encourage a general autoencoder structure model to obtain \textit{perfect} controllable disentanglement property, we propose an information-theoretic regularization with two perspectives (Fig.~\ref{fig:2}(a)):
(1) Maximize the mutual information ($I()$) between the \textit{same} attribute across latent space $\mathbb{R}^d$ and observation input space $\mathcal{X}$; and 
(2) Minimize the mutual information between the \textit{different} attributes across latent $\mathbb{R}^d$ and observation input space $\mathcal{X}$. Formally:
\vspace{-5pt}
\begin{equation}\label{Eq.1}
\vspace{-8pt}
\begin{aligned}
    &\max_{E, D}\biggl[ I(x_{\mathcal{A}_i}, E(x)_{\mathcal{A}_i}) + I(E(x)_{\mathcal{A}_i}, D(E(x))_{\mathcal{A}_i})\biggr];\\ 
    &\min_{E, D}\biggl[ I(x_{\mathcal{A}_i}, E(x)_{\mathcal{A}_j}) + I(E(x)_{\mathcal{A}_i}, D(E(x))_{\mathcal{A}_j})\biggr];\\ 
\end{aligned}
\vspace{-5pt}
\end{equation}

where $x_{\mathcal{A}_i}$ and $D(E(x))_{\mathcal{A}_i}$ represent the observation of attribute $\mathcal{A}_i$ in $\mathcal{X}$ domain (e.g., hair color in human image); $E(x)_{\mathcal{A}_i}$ represents the dimensions in $\mathbb{R}^d$ that represent attribute $\mathcal{A}_i$; $i,j\in[1..m]$ and $i\neq j$ (Fig.~\ref{fig:2}(a)).
\begin{figure}
\vspace{-15pt}
\begin{center}
\includegraphics[width=\linewidth]{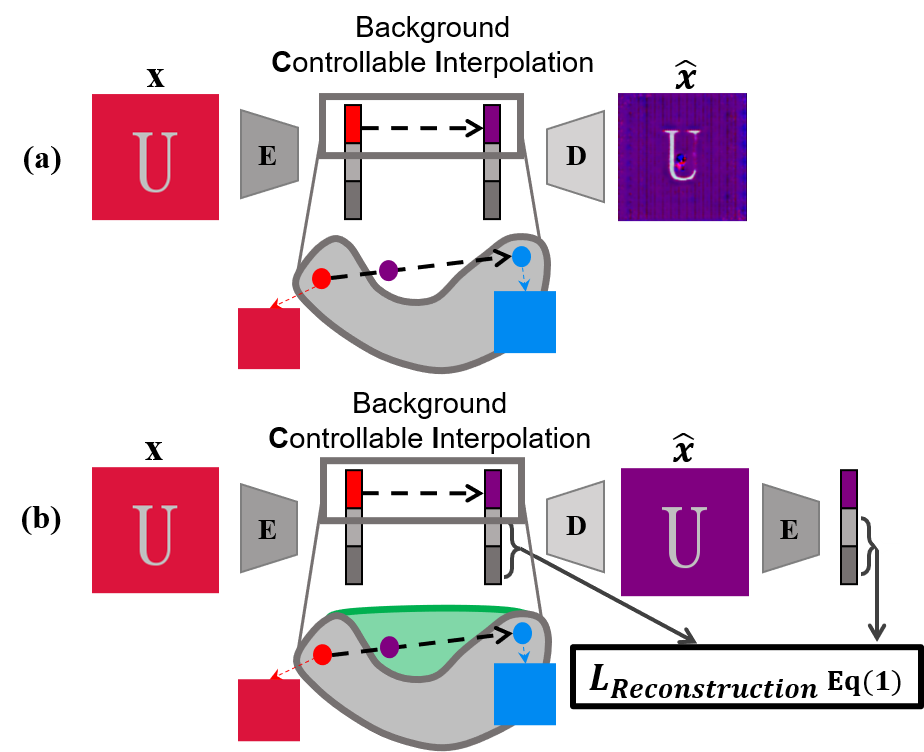}
\end{center}
\setlength{\abovecaptionskip}{-1pt}
  \caption{Interpolation in the disentangled latent space for background color of GZS-Net (a) without CIR, the latent space is not convex (purple point out of understandable gray region) and the synthesized image shows some contamination over unmodified attributes (size and foreground letter) (b) Architecture of GZS-Net + \textbf{CIR}, which encourages a more disentangled and convex latent space.}
\label{fig:3}
\vspace{-10pt}
\end{figure}
\vspace{-10pt}
\subsection{Convexity Constraint with Interpolation}
\vspace{-8pt}
\label{sec:3.2}
A convex latent space has the property that the line segment connecting any pair of points will fall within the rest of the space \cite{sainburg2018generative}. As shown in Fig.~\ref{fig:3}(a), the gray region represents the 2D projection of the latent representation of one attribute (e.g., background color) for a dataset. This distribution would be non-convex, because the purple point, though between two points in the distribution (the red and blue points, represent two background color), falls in the space that does not correspond to the data distribution. This non-convexity may cause that the projection back into the image space does not correspond to a proper semantically meaningful realistic image ($\hat{x}$ in Fig.~\ref{fig:3}(a) influence other unmodified attributes, i.e, 
size and foreground letter).
This limitation makes disentanglement vulnerable and hinders potential latent manipulation in downstream tasks. The result of Fig.~\ref{fig:4} and \ref{fig:5}in experiments illustrate this problem.\\ 
To encourage a convex data manifold, the usefulness of interpolation has been explored in the context of representation learning \cite{bengio2013better} and regularization \cite{verma2018manifold}. As is shown in Fig.~\ref{fig:1}(b), we summarize the constraint of convexity in the latent space: we use a dataset-related quality evaluation function $Q()$ to evaluate the "semantic meaningfulness" of input domain samples; a higher value means high quality and more semantic meaning. After interpolation in latent space $\mathbb{R}^d$, we want the projection back into the original space to have a high $Q()$ score. Formally:
\vspace{-3pt}
\begin{equation}\label{Eq.2}
        \max_{E, D} \biggl\{ 
         \mathbb{E}_{x_1, x_2 \in \mathcal{D}}
         \biggl[ Q(D(\alpha E(x_1) + (1-\alpha)E(x_2))) \biggr]
         \biggr\}
\vspace{-3pt}
\end{equation}
where $x_1$ and $x_2$ are two data samples and $\alpha \in [0..1]$ controls the latent code interpolation in $\mathbb{R}^d$.

The dataset-related quality evaluation function $Q()$ also has different implementations:  \cite{sainburg2018generative} utilizes additional discriminator and training adversarially on latent interpolations; \cite{berthelot2018understanding} uses a critic network as a surrogate which tries to recover the mixing coefficient from interpolated data.

\vspace{-10pt}
\subsection{CIR: encourage both C-Dis-RL and Convexity}
\vspace{-8pt}
\label{sec:3.3}



Our goal is to encourage a controllable disentangled representation, and, for each semantic attribute-related latent dimension, the created space should be as convex as possible. Specifically, we want to optimize both controllable disentanglement (Eq.~\ref{Eq.1}) and convexity (Eq.~\ref{Eq.2}) for each semantic attribute. In practice, each mutual information term in Eq.~\ref{Eq.1} is hard to optimize directly as it requires access to the posterior. Most of the current methods use approximation to obtain the lower bound for optimizing the maximum \cite{chen2016infogan, belghazi2018mutual} or upper bound for optimizing minimum \cite{kingma2014autoencoding}. However, it is hard to approximate so many  ($2m(m-1)+2m$) different mutual information terms in Eq.~\ref{Eq.1}) simultaneously, not to mention considering the convexity of $m$ latent space (Eq.~\ref{Eq.2}) as well. 
To optimize them together, we propose to use a controllable disentanglement constraint to help the optimization of convexity and in turn, use convexity constraint to help a more robust optimization of the controllable disentanglement. In other words, we create a positive loop between controllable disentanglement and convexity, to help each other. Specifically, as shown in Fig.~\ref{fig:1}(c), we propose a simple yet efficient regularization method, Controllable Interpolation Regularization (CIR), which consists of two main modules: a Controllable Interpolation (CI) module and a Reuse Encoder Regularization (RER) module.
It works as follows: an input sample $x$ goes through $E$ to obtain latent code $z=E(x)$. Because our goal is controllable disentanglement, on each iteration we only focus on one attribute. CI module first selects one attribute $\mathcal{A}_i$ among all $m$ attributes, and then interpolates along the $\mathcal{A}_i$ related latent space in $z$ while preserving the other unselected attributes, yielding $z_{\mathcal{A}_i}$. After $D$ translates the interpolated latent $z_{\mathcal{A}_i}$ back to image space, the RER module takes $D(z_{\mathcal{A}_i})$ as input and reuses the encoder to get the latent representation $z^{re}_{\mathcal{A}_i} = E(D(z_{\mathcal{A}_i}))$. RER then adds a reconstruction loss on the \textit{unmodified latent space} as a regularization:
\begin{equation}\label{Eq.3}
	L_\textrm{reg}=|| z_{-\mathcal{A}_i} -  z^{re}_{-\mathcal{A}_i} ||_{l1}
\end{equation}
where $z_{-\mathcal{A}_i}$ and $z^{re}_{-\mathcal{A}_i}$ denote the all latent dimensions of $z_{\mathcal{A}_i}$ and $z^{re}_{\mathcal{A}_i}$ respectively, except those that represent the modified attribute $\mathcal{A}_i$. 
Eq.~\ref{Eq.3} explicitly optimizes Eq.~\ref{Eq.1}: in each iteration, if the modified latent region $z_{\mathcal{A}_i}$ only influences the expression of $x_{\mathcal{A}_i}$, then, after reusing $E$, the unmodified region in $E(D(z_{\mathcal{A}_i}))$ should remain as is (min E, D in Eq.~\ref{Eq.1}). On the one hand, for those unselected attributes, their information should be preserved in the whole process (max E, D in Eq.~\ref{Eq.1}).  Eq.~\ref{Eq.3} also implicitly optimizes Eq.~\ref{Eq.2}: if the interpolated latent code is not 'understandable' by $E$ and $D$, the RER module does not work and the $L_\textrm{reg}$ would be large. Fig.~\ref{fig:2} (a) and (b) abstractly demonstrate the latent space convexity difference before and after adding CIR to GZS-Net \cite{ge2020zero}. 
Convexity and disentanglement are dual tasks in the sense that one can help enhance the other's performance. On the other hand, the reconstruction loss towards \textit{perfect} controllable disentanglement implicitly encourages a convex attribute latent space; The more convex the latent space, the more semantically meaningful samples synthesized by interpolation will help the optimization of controllable disentanglement, which encourages a more robust C-Dis-RL. 
From the perspectives of loss function and optimization, 
if the reconstruction loss could decrease to zero for a given dataset augmented by many interpolated samples, 
then perfect disentanglement and convexification are achieved. That is, CIR forces, in the limit of infinite interpolated samples, the disentangled latent representation of every attribute to be $convex$, where every interpolation along every attribute is guaranteed to be meaningful.

\vspace{-10pt}
\section{Qualitative Experiments}
\vspace{-8pt}
\label{sec:4}

We qualitatively evaluated our CIR as a general module and merged it into three baseline models on three different tasks: multiple face attributes transfer with ELEGANT \cite{xiao2018elegant} (Sec.~\ref{exp:4.1}), cross modality image translation with I2I-Dis \cite{lee2018diverse} (Sec.~\ref{exp:4.2}) and zero-shot synthesis with GZS-Net \cite{ge2020zero} (Sec.~\ref{exp:4.3}).  CIR encourages a better disentanglement and convexity in their latent space to further improve their performance.

\begin{figure}
\vspace{-30pt}
\begin{center}
\includegraphics[width=\linewidth]{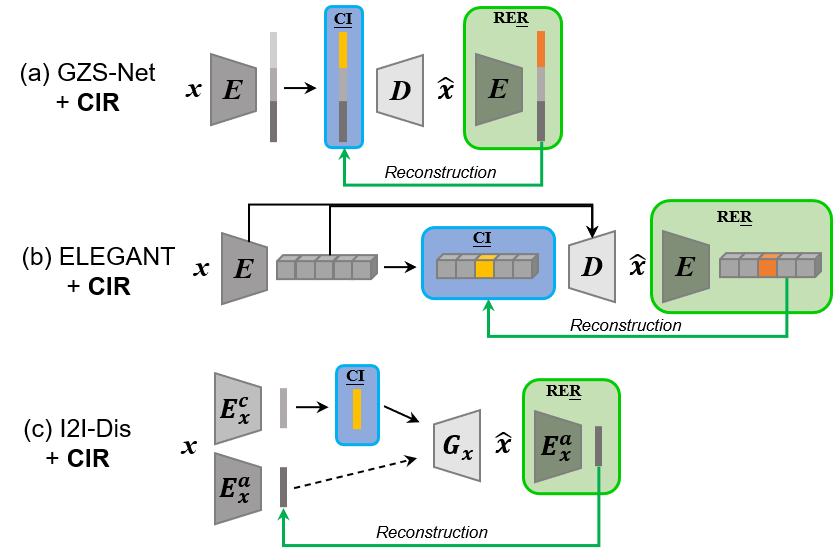}
\end{center}
\setlength{\abovecaptionskip}{-1pt}
  \caption{CIR consists of a Controllable Interpolation (CI; shown in blue) module and a Reuse Encoder Regularization (RER; green) module. (a-c) \textbf{CIR} compatible to different models. (a) GZS-Net \cite{ge2020zero} + \textbf{CIR} (b) ELEGANT \cite{xiao2018elegant} + \textbf{CIR}. (c) I2I-Dis \cite{lee2018diverse} + \textbf{CIR}. Grey components are the baseline methods. }
\label{fig:4}
\vspace{-15pt}
\end{figure}

\vspace{-12pt}
\subsection{\textbf{CIR} boosts multiple face attributes transfer}
\vspace{-8pt}
\label{exp:4.1}

\begin{figure}[t]
\vspace{-35pt}
  \centering
  \includegraphics[width=\linewidth]{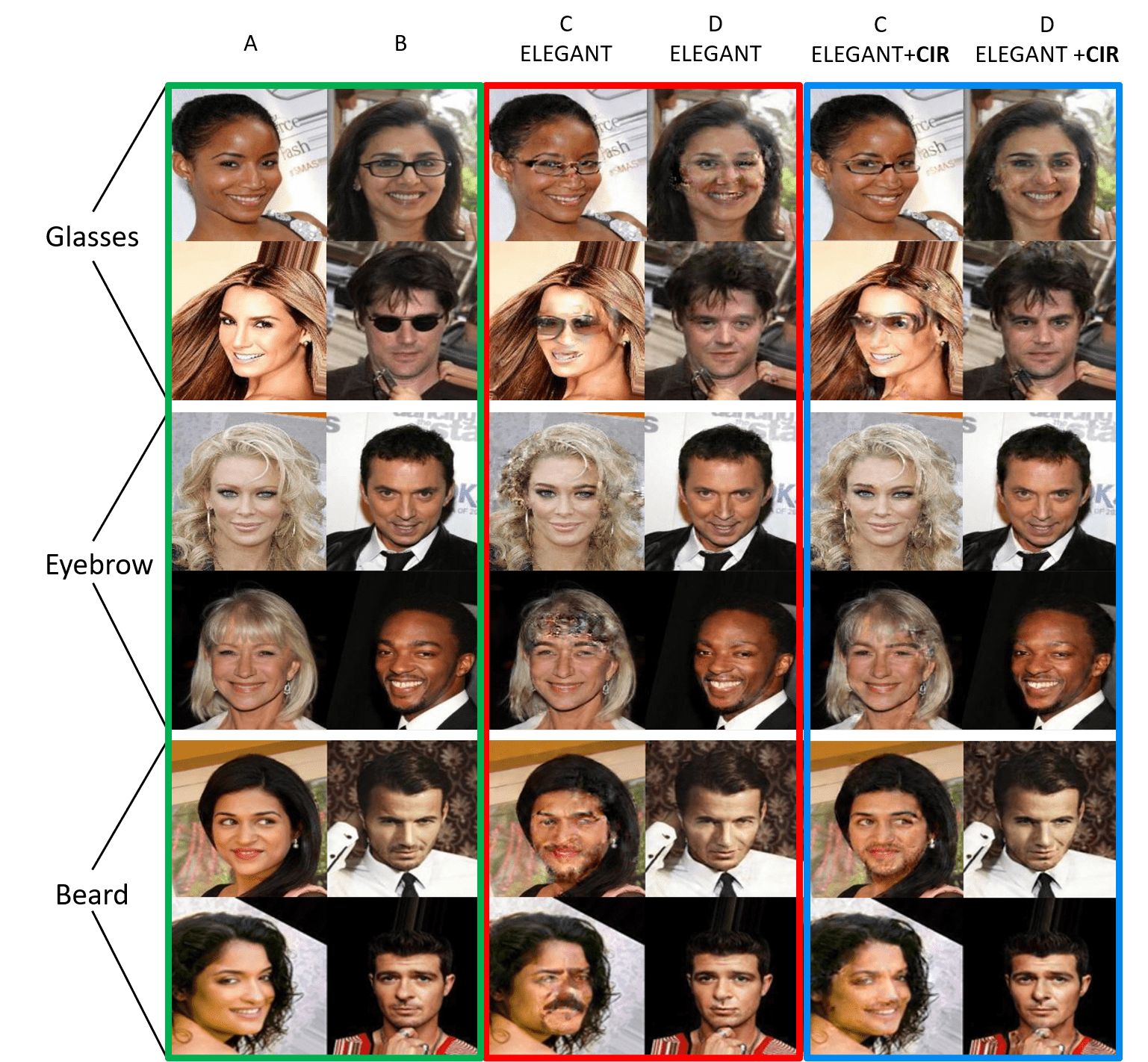}
  \caption{
  ELEGANT + \textbf{CIR} performance (task 1) for two images face attribute transfer (inputs: A,B ; outputs: C,D).}
  \label{fig:5}
  \vspace{-10pt}
\end{figure}

\begin{figure}[t]
\vspace{-25pt}
  \centering
  \includegraphics[width=\linewidth]{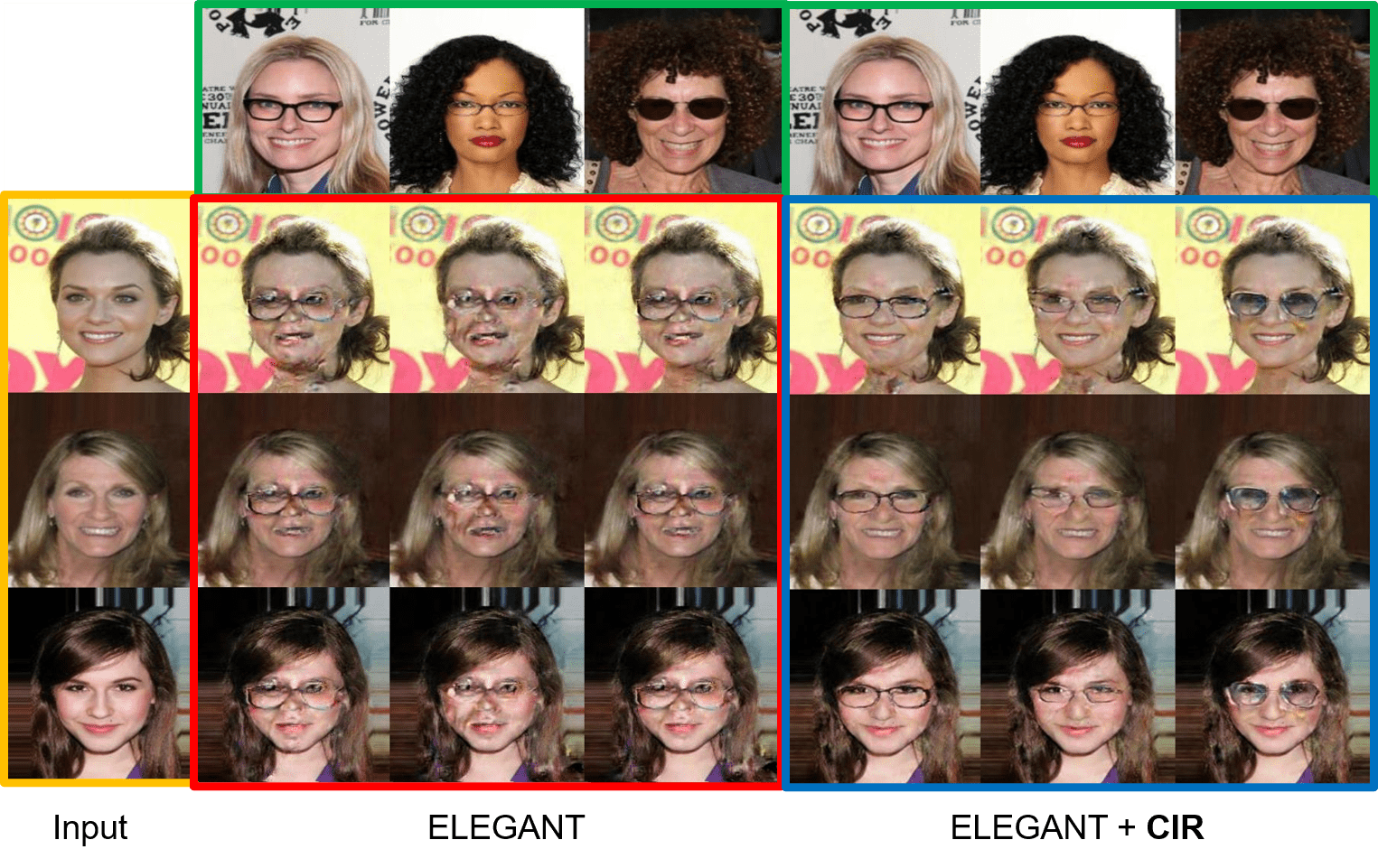}
  \caption{
  ELEGANT + \textbf{CIR} performance (task 2) for face generation by exemplars: Input image (orange) should be modified as different face images with the same style of the Eyeglasses attribute in the reference images (green).}
  \label{fig:6}
\vspace{-15pt}
\end{figure}

We conduct the same face attribute transfer tasks as in ELEGANT \cite{xiao2018elegant} paper with \textit{CelebA} \cite{liu2015deep}. \textbf{Task 1}: taking two face images with the opposite attribute as input and generate new face images which exactly transfer the opposite attribute between each other (Fig.~\ref{fig:5}).  \textbf{Task 2}: generate different face images with the same style of the attribute in the reference images (Fig.~\ref{fig:6}). Both of the two tasks require a robust controllable disentangled latent space to swap attributes of interest to synthesize new images and the convexity of latent space influences image quality. 

Fig.~\ref{fig:4}(b) shows the high-level structure about how CIR (blue and green block) compatible to ELEGANT (grey). 
ELEGANT adopts a U-Net \cite{ronneberger2015u} structure (autoencoder) to generate high-resolution images with exemplars. In this way, the output of the encoder is the latent code of disentangled attributes and the context information is contained in the output of the intermediary layer of the encoder. ELEGANT adopts an iterative training strategy: training the model with respect to a particular attribute each time. We use the same training strategy but adding our regularization loss term. As shown in Fig.~\ref{fig:4} (b), to encourage the disentanglement and convexity of attribute $\mathcal{A}_i$, CIR interpolates $\mathcal{A}_i$-related dimensions in latent code (yellow) and constrains the other latent dimensions to remain unchanged after $D$ and reused $E$. Specifically, when training ELEGANT about the $\mathcal{A}_i$ attribute \textbf{Eyeglasses} at a given iteration, we obtain the latent code $zA = E(A)$ and $zB = E(B)$ with $E$ for each pair of images $A$ and $B$ with opposite $\mathcal{A}_i$ attribute value. The disentangled latent code is partitioned into $z_{+\mathcal{A}_i}$ for latent dimensions related to $\mathcal{A}_i$, and $z_{-\mathcal{A}_i}$ for unrelated dimensions. We interpolate in $z_{+\mathcal{A}_i}$ with $zA$ and $zB$ while keeping the other dimensions $z_{-\mathcal{A}_i}$ as is to obtain interpolated latent code $zA_{\mathcal{A}_i}$ and $zB_{\mathcal{A}_i}$. After $D$ and reuse $E$, we get the reconstructed latent representation $zA_{\mathcal{A}_i}^{re} = E(D(zA_{\mathcal{A}_i}, zA))$ and $zB_{\mathcal{A}_i}^{re} = E(D(zB_{\mathcal{A}_i}, zB))$. The reconstruction loss as a regularization is (an instantiation of Eq.~\ref{Eq.3}):
\begin{equation}\label{Eq.4}
	L_\textrm{reg}=|| zA_{-\mathcal{A}_i} -  zA^{re}_{-\mathcal{A}_i} ||_{l2} + || zB_{-\mathcal{A}_i} -  zB^{re}_{-\mathcal{A}_i} ||_{l2}
\end{equation}
The overall generative loss of ELEGANT + CIR is:
\begin{equation}\label{Eq.5}
	\mathcal{L}(G)=L_\textrm{reconstruction} + L_\textrm{adv} + \bm{\lambda_\textrm{CIR}L_\textrm{reg}}
\end{equation}
where $L_\textrm{reconstruction}$ and $L_\textrm{adv}$ are ELEGANT original loss terms, $\lambda_\textrm{CIR} > 0$ control the relative importance of the loss terms. 
we keep the discriminative loss. (More network architecture and training details are in Supplementary)

Fig.~\ref{fig:5} shows the task 1 performance on two images face attribute transfer. Take Eyeglasses as an example attribute to swap: A,B are input, the output C and D should keep all other attributes unmodified except for swapping the Eyeglasses. ELEGANT generated C and D have artifacts in Eyeglasses-unrelated regions, which means ELEGANT cannot disentangle well in latent space. After adding CIR, the generated C and D better preserve the irrelevant regions during face attribute transfer, which demonstrates that CIR helps encourage a more convex and disentangled latent space. The Eyebrow and Beard attribute results also show the improvement from CIR.
Fig.~\ref{fig:6} shows the task 2 performance on face image generation by exemplars. Input image (orange) should be modified as different face images with the same style of the Eyeglasses attribute in the reference images (green).
ELEGANT generated new images with artifacts in Eyeglasses-unrelated regions that cannot disentangle well. Synthesis is also inferior in the glasses region, which we posit is due to non-convexity in the eyeglass-related latent space. With the help of CIR, the generated images improve both Eyeglass quality and irrelevant region preservation.

\vspace{-12pt}
\subsection{\textbf{CIR} boosts cross modality image translation}
\begin{figure}[t]
\vspace{-25pt}
  \centering
  \includegraphics[width=\linewidth]{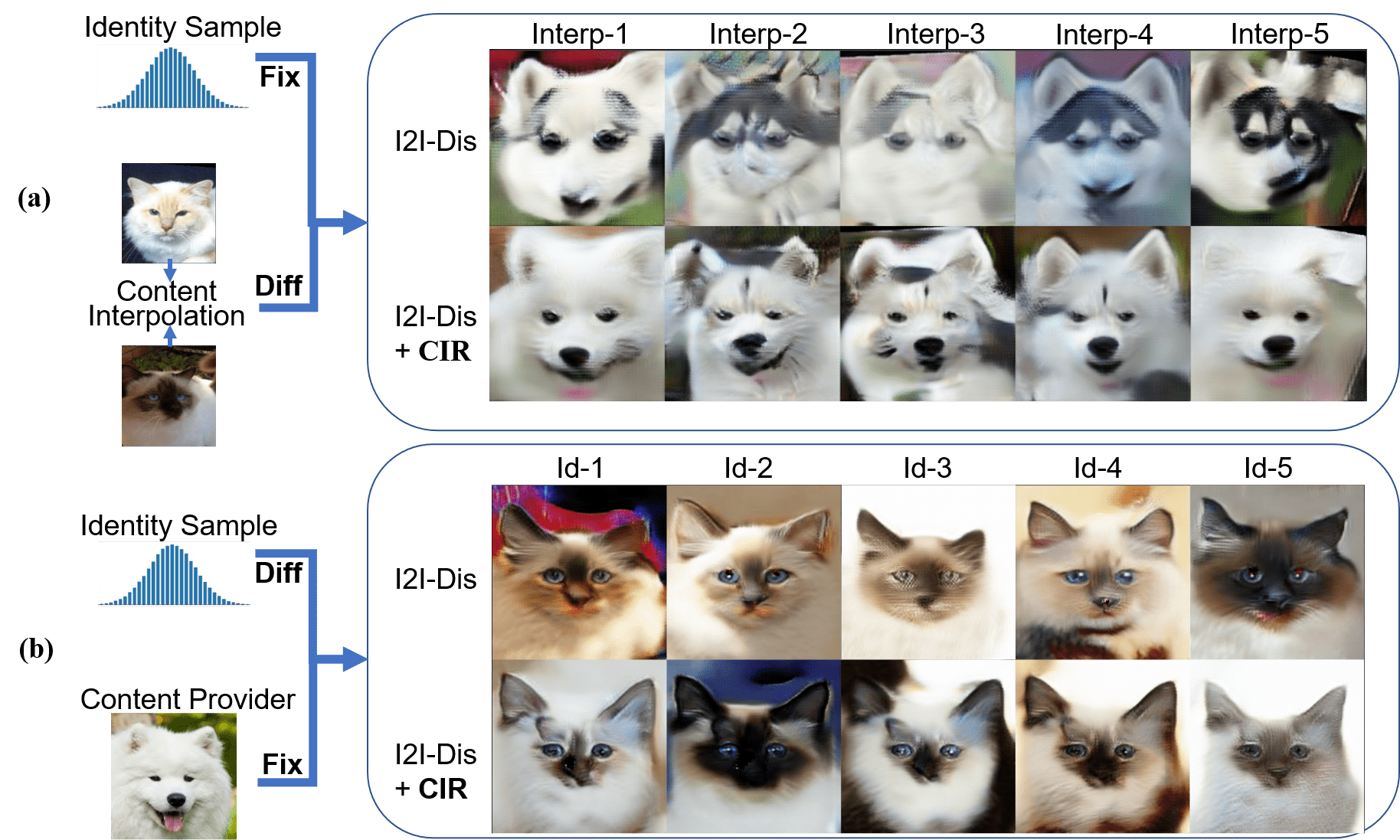}
  \caption{
  I2I-Dis + \textbf{CIR} performance of diverse image-to-image translation. (a) For any dog image sample, create several interpolated images with content (here, pose, ear orientation, etc) in between that of the two reference cat images. (b) For several cat identity samples, synthesize images with that cat's identity but the content (pose, etc) of the reference dog image.}
  \label{fig:7}
\vspace{-15pt}
\end{figure}

\vspace{-8pt}
\label{exp:4.2}
We conduct the same image-to-image translation task as in I2I-Dis \citep{lee2018diverse} paper with \textit{cat2dog} dataset \citep{lee2018diverse}. Fig.~\ref{fig:4}(c) shows the high-level structure about how CIR (blue and green block) compatible to I2I-Dis (grey). There are two image domains $\mathcal{X}$ (cat) and $\mathcal{Y}$ (dog), I2I-Dis embeds input images onto a shared content space $\mathcal{C}$ with specific encoders ($E_{\mathcal{X}}^{c}$ and $E_{\mathcal{Y}}^{c}$), and domain-specific attribute spaces $\mathcal{A}_{\mathcal{X}}$ and $\mathcal{A}_{\mathcal{Y}}$ with specific encoders ($E_{\mathcal{X}}^{a}$ and $E_{\mathcal{Y}}^{a}$) respectively. After that, new images can be synthesized by transferring the shared content attribute cross-domain (between cat and dog), such as generating unseen dogs with the same content attribute value (pose and outline) as the reference cat (Fig.~\ref{fig:7}). Domain-specific attribute $\mathcal{A}_{\mathcal{X}}$ and $\mathcal{A}_{\mathcal{Y}}$ already been constraint by adding a KL-Divergence loss with Gaussian distribution; thus, we can freely sample in Gaussian for synthesis. The shared content space $\mathcal{C}$ could be encouraged as a more convex and disentangled space by CIR. 

We use the same network architecture and training strategy as I2I-Dis  
except for adding our regularization loss term. As shown in Fig.~\ref{fig:4} (c), during each training iteration, a cat image $x$ and a dog image $y$ go through corresponding encoders and each of them produce latent codes of domain ($zx_a = E_{\mathcal{X}}^{a}(x)$, $zy_a = E_{\mathcal{Y}}^{a}(y)$ ) and content ($zx_c = E_{\mathcal{X}}^{c}(x)$, $zy_c = E_{\mathcal{Y}}^{c}(y)$). 
Then an interpolated content attribute latent code (yellow) $zxy_c$ (between $zx_c$ and $zy_c$) concatenates with the domain attribute latent code of cat image $zx_a$ and dog image $zy_a$ respectively and forms two new latent codes, and decoders turns them into new images $u = G_{\mathcal{X}}(zx_a, zxy_c)$, $v = G_{\mathcal{Y}}(zy_a, zxy_c)$. 
To encourage the disentanglement and convexity of the content attribute, we reuse $E_{\mathcal{X}}^{a}$ and $E_{\mathcal{Y}}^{a}$ to get the reconstructed domain attribute latent representations $zx_a^{re} = E_{\mathcal{X}}^{a}(u)$, $zy_a^{re} = E_{\mathcal{Y}}^{a}(v)$ and add the reconstruction loss as a regularization (an instantiation of Eq.~\ref{Eq.3}):
\begin{equation}\label{Eq.4}
	L_\textrm{reg}=|| zx_{a}^{re} -  zx_{a} ||_{l1} + || zy_{a}^{re} -  zy_{a} ||_{l1}
\end{equation}
The overall loss of I2I-Dis + CIR is 
\begin{equation}
    \footnotesize
    \begin{split}
        \mathcal{L} = & \lambda_{adv}^{content}L_{adv}^{c} + \lambda_{1}^{cc}L_{1}^{cc} +  \lambda_{adv}^{domain}L_{adv}^{domain} + \\ &\lambda_{1}^{recon}L_{1}^{recon} +
	    \lambda_{1}^{latent}L_{1}^{latent} + \lambda_{KL}L_{KL} + 
	    \bm{\lambda_\textrm{CIR}L_\textrm{reg}}
    \label{Eq.5}
    \end{split}
\end{equation}
where content and domain adversarial loss $L_{adv}^{c}$ $L_{adv}^{domain}$, cross-cycle consistency loss $L_{1}^{cc}$, self-reconstruction loss $L_{1}^{recon}$, latent regression loss $L_{1}^{latent}$ and KL loss $L_{KL}$ are I2I-Dis original loss terms, $\lambda > 0$ control the relative importance of the loss terms. (More details in Supplementary).

Fig.~\ref{fig:7} shows the image-to-image translation performance. (a) We fix the identity (domain) latent code and change the content latent code by interpolation; generated images should keep the domain attribute (belong to the same dog). I2I-Dis generated dog images have artifacts, which means the non-convex latent space cannot 'understand' the interpolated content code. After adding our CIR, the generated images have both better image quality and consistency of the same identity. (b) We fix the content latent code and change the identity by sampling; generated images should keep the same content attribute (pose and outline). Cat images generated by I2I-Dis have large pose variance (contain both left and right pose), and large face outline variance (ear positions and sizes). After adding our CIR, the generated images have smaller pose and outline variance. (More results in Supplementary)

\vspace{-12pt}
\subsection{\textbf{CIR} boosts zero-shot synthesis}
\vspace{-10pt}
\label{exp:4.3}
We use the same architecture of autoencoders as GZS-Net \cite{ge2020zero} and \textit{Fonts} dataset \cite{ge2020zero}. Fig.~\ref{fig:4}(a) shows the high-level structure about how CIR (blue and green block) compatible to GZS-Net (grey). The latent feature after encoder $E$ is a 100-dim vector, and each of the five \textit{Fonts}  attributes (content, size, font color, background color, font) covers 20-dim. The decoder $D$, symmetric to $E$, takes the 100-dim vector as input and outputs a synthesized sample. We use the same Group-Supervised learning training strategy as GZS-Net except for adding our regularization loss term Eq.~\ref{Eq.1}, which is exactly the same as the one described in Sec.~\ref{sec:3.3} and Fig.~\ref{fig:3} (b).
Besides the reconstruction loss $L_\textrm{r}$, swap reconstruction loss $L_\textrm{sr}$ and cycle swap reconstruction loss $L_\textrm{csr}$ which are same as GSL, we add a regularization reconstruction loss $L_\textrm{reg}$. The total loss function is:
\begin{equation}\label{Eq.6}
\vspace{-3pt}
	\mathcal{L}(E, D)=L_\textrm{r} + \lambda_\textrm{sr}L_\textrm{sr} + \lambda_\textrm{csr}L_\textrm{csr} + \bm{\lambda_\textrm{CIR}L_\textrm{reg}}
\vspace{-3pt}
\end{equation}
where $\lambda_\textrm{sr}, \lambda_\textrm{csr}, \lambda_\textrm{CIR} > 0$ control the relative importance of the loss terms. 


\begin{figure}
\vspace{-25pt}
  \centering
  \includegraphics[width=\linewidth]{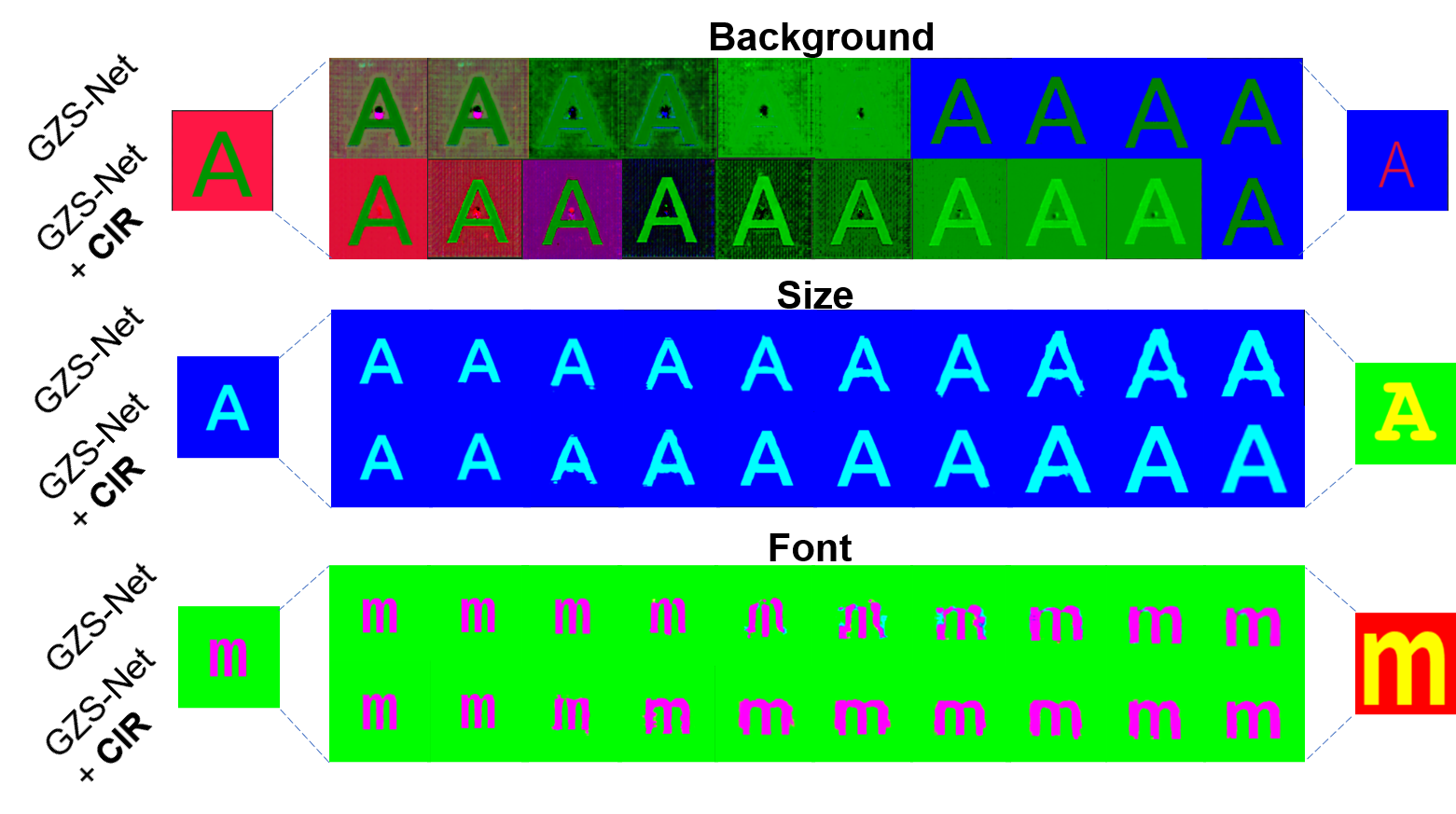}
  \caption{
  GZS-Net + \textbf{CIR} performance of interpolation-based attribute controllable synthesis. Top: Interpolation in the latent space of background color. Middle: interpolation of letter size. Bottom: Interpolation of font style. In all three cases, CIR provides both better disentanglement (attributes other than the interpolated one do not change as much) and higher interpolation quality (interpolated attributes show fewer artifacts).}
  \label{fig:8}
\vspace{-10pt}
\end{figure}

Fig.~\ref{fig:8} shows the interpolation-based controllable synthesis performance on background, size, and font attributes. 
Take background interpolation synthesis as an example: we obtain background latent codes by interpolating between the left and right images, and each of them concatenates with the unselected 80-dim latent code from the left image. Generated images should keep all other attributes unmodified except for the background. GZS-Net generated images have artifacts in background-unrelated regions, i.e., GZS-Net cannot disentangle well in latent space. After adding our CIR, the generated images better preserve the irrelevant areas during synthesis. The size and font attribute results also show improvement from CIR. (More results in Supplementary).

\vspace{-12pt}
\section{Quantitative Experiments}
\vspace{-8pt}
\label{sec:5}
We conduct five quantitative experiments to evaluate the performance of CIR on latent disentanglement and convexity.
\noindent{\bf Controllable Disentanglement Evaluation by Attribute Co-prediction.}
Can latent features of one attribute predict the attribute value? Can they also predict values for other attributes?
Under \textit{perfect} controllable disentanglement, we should answer \textit{always} for the first and \textit{never} for the second.
We quantitatively assess disentanglement by 
calculating a model-based confusion matrix between attributes. We evaluate GZS-Net \cite{ge2020zero} + \textbf{CIR} with the \textit{Fonts} \cite{ge2020zero} dataset (latent of ELEGANT and I2I-Dis are not suitable). Each image in \textit{Fonts} contains an alphabet letter rendered using 5 independent attributes: content (52 classes), size (3), font color (10), background color (10), and font (100). 
We take the test examples and split them 80:20 for $\textrm{train}_\textrm{DR}$:$\textrm{test}_\textrm{DR}$. For each attribute pair $j, r \in [1..m] \times [1..m]$, we train a classifier (3 layer MLP) from $g_j$ of $\textrm{train}_\textrm{DR}$ to the attribute values of $r$, then obtain the accuracy of each attribute by testing with $g_j$ of $\textrm{test}_\textrm{DR}$.
Fig.~\ref{fig:9} compares how well features of each attribute (row) can predict an attribute value (column): perfect should be as close as possible to Identity matrix, with off-diagonal entries close to random (i.e., 1 / $|\mathcal{A}_r|$).
The off-diagonal values of GZS-Net show the limitation of disentanglement performance; with CIR's help, the co-prediction value shows a better disentanglement.


\begin{figure}
\vspace{-25pt}
  \centering
  \includegraphics[width=\linewidth]{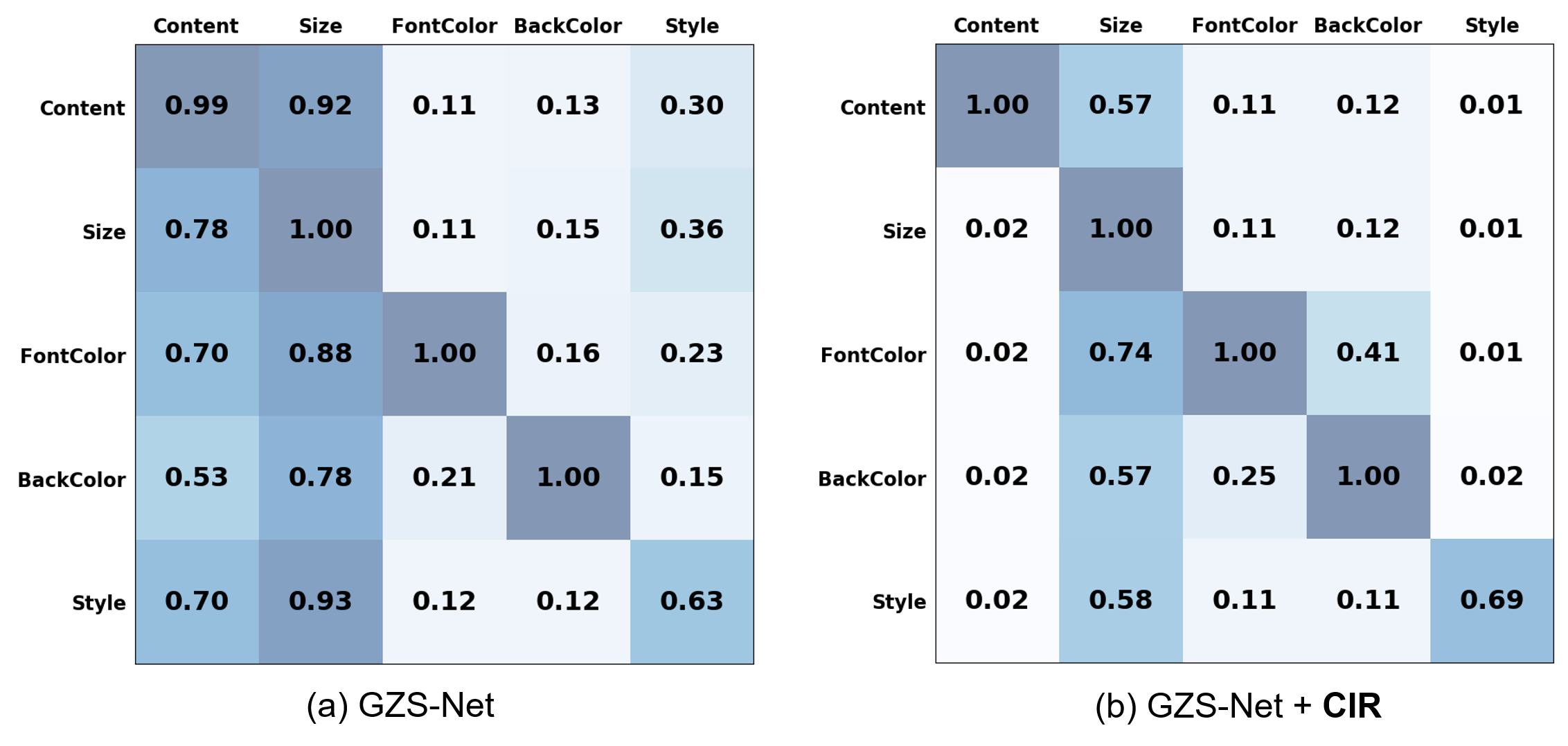}
  \caption{
  Disentangled representation analysis. (a) Baseline GZS-Net. (b) With CIR, off-diagonal elements (entanglement across attributes) are reduced.}
  \label{fig:9}
\vspace{-15pt}
\end{figure}
\begin{figure}
\vspace{-20pt}
  \centering
  \includegraphics[width=\linewidth]{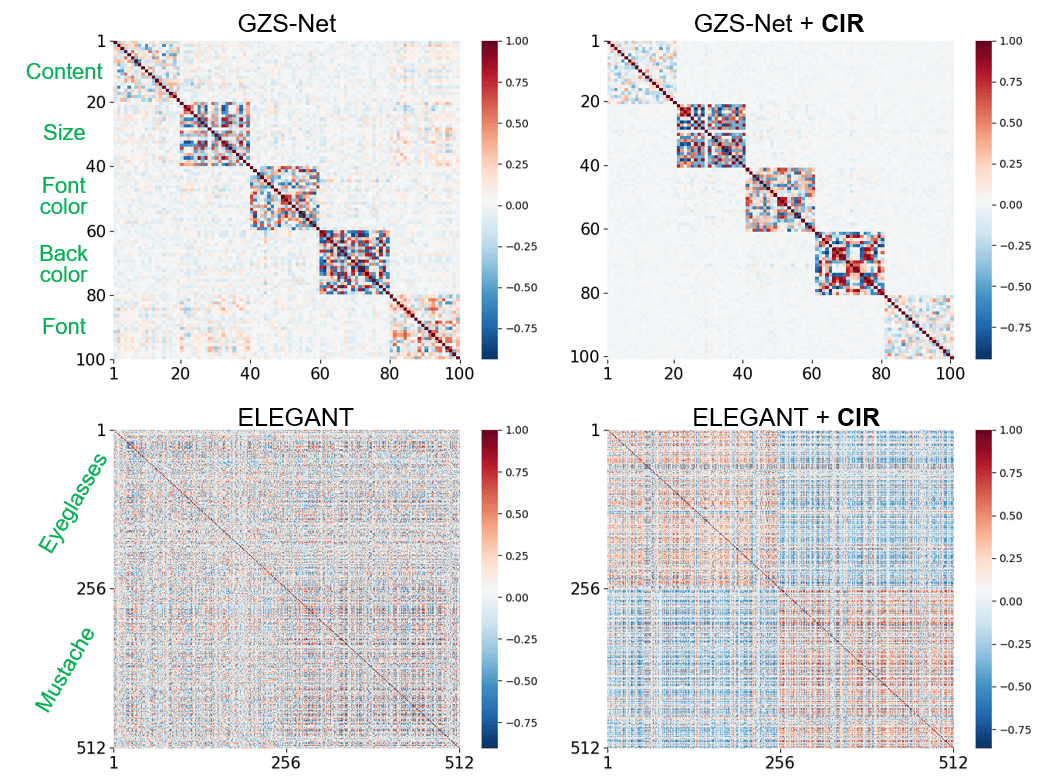}
  \caption{
  Disentanglement Evaluation by Correlation Coefficient. Intra-attribute correlation increases with CIR (GZS-Net (top): $7.2\%$, ELEGANT (bottom): $3.2\%$) while inter-attribute decreases (GZS-Net: $60.9\%$, ELEGANT: $3.1\%$).}
  \label{fig:10}
\vspace{-5pt}
\end{figure}

\vspace{-1pt}
\noindent{\bf Controllable Disentanglement Evaluation by Correlation Coefficient.}
For each method, we collect 10,000 images from the corresponding dataset (ELEGANT \citep{xiao2018elegant} with \textit{CelebA} \citep{liu2015deep}, GZS-Net with \textit{Fonts} \citep{ge2020zero}) and obtain 10,000 latent codes by $E$s. We calculate the correlation coefficient matrix between dimensions in latent space. A near-perfect disentanglement should yield high intra-attribute correlation but low inter-attribute correlation. ELEGANT disentangles two attributes: eyeglasses and mustache, each of which covers 256-dimensions. GZS-Net disentangles five attributes: content, size, font color, background color, and font; each covers 20-dimensions. Fig.~\ref{fig:10} shows that CIR improves the disentanglement in latent space, as demonstrated by higher intra-attribute and lower inter-attribute correlations (More details in Suppl.). 


\noindent{\bf Convexity Evaluation with Image Quality Score.}
To evaluate the overall convexity in latent space, we use an image quality classifier to evaluate the quality of images generated by interpolating in latent space. 
We train a specific image quality classifier for each baseline algorithm and corresponding dataset. Take ELEGANT as an instance: To train a classifier for ELEGANT and ELEGANT + CIR, we use 3000 \textit{CelebA} original images as positive, high-quality images. To collect negative images, we first randomly interpolate the latent space of both ELEGANT and ELEGANT + CIR and generate interpolated images for negative low-quality images; then, we manually select 3000 low-quality images (artifact, non-sense, fuzzy ...) and form a 6000 images training set. After training an image quality classifier, we test it on 1500 images generated by interpolation-based attribute controllable synthesis as Exp.~\ref{exp:4.1}. Table~\ref{table-1} shows the average probability of high-quality images (higher is better). The training and testing for I2I-Dis (+ CIR) and GZS-Net (+ CIR) are similar.

\begin{table}
  \scriptsize
  \setlength{\abovecaptionskip}{3pt}
\caption{Convexity Evaluation with Image Quality Score}
\label{table-1}
\begin{tabular}{c|c|c|c}
\hline
Algorithms & Train images & Test images & High quality probability \\
\hline
ElEGANT & \multirow{2}{*}{6000} & \multirow{2}{*}{1500} & 12\% \\

ElEGANT + \textbf{CIR} & &  & \textbf{60\%} \\
\hline

I2I-Dis & \multirow{2}{*}{1500} & \multirow{2}{*}{1500} & 18\% \\

I2I-Dis + \textbf{CIR} &  &  & \textbf{33\%} \\
\hline
GZS-Net & \multirow{2}{*}{6000} & \multirow{2}{*}{1000} & 13\% \\

GZS-Net + \textbf{CIR} &  &  & \textbf{40\%} \\
\hline
\end{tabular}
\vspace{-15pt}
\end{table}

\noindent{\bf \textit{Perfect} Disentanglement Property Evaluation.}
As we defined in Sec.~\ref{sec:3.1}, \textit{Perfect} disentanglement property can be evaluated by the difference of the unmodified attribute related dimensions in $\mathbb{R}^d$ after modifying a specific attribute $\mathcal{A}_i$ in image space. For the two methods in each column (Table~\ref{table-2}) and corresponding datasets, 
we modify one attribute value $\mathcal{A}_i$ of each input $x$ and get $\hat{x}$, then obtain latent codes ($z = E(x)$, $\hat{z} = E(\hat{x})$) with two methods' encoders respectively. After we normalized the latent codes from two methods into the same scale,
we calculate the Mean Square Error (MSE) of the unmodified region $MSE(z_{-\mathcal{A}_i}, \hat{z}_{-\mathcal{A}_i})$ between $z$ and $\hat{z}$ (lower is better).
Table~\ref{table-2} shows that after adding CIR, we obtain a lower MSE, which means CIR encourages a better disentangled latent space.


\begin{table}
\vspace{-20pt}
  \scriptsize
  \setlength{\abovecaptionskip}{3pt}
\caption{\textit{Perfect} Disentanglement Property Evaluation}
\label{table-2}
\begin{tabular}{c|c|c|c}
\hline
Algorithms & ElEGANT & I2I-Dis & GZS-Net \\
\hline
MSE  & 1.9 & 1.8  & 3.42  \\
\hline
Algorithms & ElEGANT + \textbf{CIR} &  I2I-Dis + \textbf{CIR} &  GZS-Net + \textbf{CIR} \\
\hline
MSE & \textbf{0.38} & \textbf{0.1} & \textbf{0.27}   \\
\hline
\end{tabular}
\vspace{-8pt}
\end{table}

\noindent{\bf C-Dis Evaluation with Perceptual Path Length Metric}
We use a method similar to the perceptual path length metric proposed by StyleGAN \citep{karras2019style}, which measure the difference between consecutive images (their VGG16 embeddings) when interpolating between two random inputs. We subdivide a latent space interpolation path into linear segments. The definition of total perceptual length of this segmented path is the sum
of perceptual differences over each segment. In our experiment, we use a small subdivision epsilon $\epsilon = 10^{-4}$ and linear interpolation (lerp). Thus, the average perceptual path length in latent space ${\mathcal{Z}}$ is
$$l_{\mathcal{Z}}=\mathbb{E}\left[\frac{1}{\epsilon^{2}} d\left(G\left(\operatorname{lerp}\left(\mathbf{z}_{1}, \mathbf{z}_{2} ; t\right)\right), G\left(\operatorname{lerp}\left(\mathbf{z}_{1}, \mathbf{z}_{2} ; t+\epsilon\right)\right)\right)\right]$$
Where $\mathbf{z}_{1}, \mathbf{z}_{2}$ is the start point and the end point in latent space. $G$ can be a decoder in Auto-encoder or generator in a GAN-based model. $t\sim U(0,1)$. $d$ is the distance in VGG16 embeddings. Our results can be seen in table.~\ref{table:3} where CIR improves the latent disentanglement.

\begin{table}
\centering
    \caption{Disentanglement Evaluation with StyleGAN Perceptual Path Length Metric. Lower difference is better.}
  \scriptsize
    \begin{tabular}{c|c|c|c}
    \hline
    I2I-Dis  &  I2I-Dis + \textbf{CIR} & ElEGANT & ElEGANT + \textbf{CIR} \\
    29 & 21  & 1.23 & 0.68  \\
    \hline
    \end{tabular}

    \label{table:3}
\end{table}

\vspace{-10pt}
\section{Conclusion}
\vspace{-5pt}

We proposed a general disentanglement module, Controllable Interpolation Regularization (CIR), compatible with different algorithms to encourage more convex and robust disentangled representation learning. We show the performance of CIR with three baseline methods ELEGANT, I2I-Dis, and GZE-Net. CIR first conducts controllable interpolation in latent space and then 'reuses' the encoder to form an explicit disentanglement constraint. Qualitative and quantitative experiments show that CIR improves baseline methods performance on different controllable synthesis tasks: face attribute transfer, diverse image-to-image transfer, and zero-shot image synthesis with different datasets: \textit{CelebA}, \textit{cat2dog} and \textit{Fonts} respectively. 



\section{Acknowledgment}

This work was supported by C-BRIC (one of six centers in JUMP, a
Semiconductor Research Corporation (SRC) program sponsored by DARPA),
DARPA (HR00112190134), the Army Research Office (W911NF2020053), and the
Intel and CISCO Corporations. The authors affirm that the views
expressed herein are solely their own, and do not represent the views of
the United States government or any agency thereof.

\bibliography{uai2022-template}

\begin{thebibliography}{26}
\providecommand{\natexlab}[1]{#1}
\providecommand{\url}[1]{\texttt{#1}}
\expandafter\ifx\csname urlstyle\endcsname\relax
  \providecommand{\doi}[1]{doi: #1}\else
  \providecommand{\doi}{doi: \begingroup \urlstyle{rm}\Url}\fi

\bibitem[Belghazi et~al.(2018)Belghazi, Baratin, Rajeshwar, Ozair, Bengio,
  Courville, and Hjelm]{belghazi2018mutual}
Mohamed~Ishmael Belghazi, Aristide Baratin, Sai Rajeshwar, Sherjil Ozair,
  Yoshua Bengio, Aaron Courville, and Devon Hjelm.
\newblock Mutual information neural estimation.
\newblock In \emph{International Conference on Machine Learning}, pages
  531--540. PMLR, 2018.

\bibitem[Bengio et~al.(2013)Bengio, Mesnil, Dauphin, and
  Rifai]{bengio2013better}
Yoshua Bengio, Gr{\'e}goire Mesnil, Yann Dauphin, and Salah Rifai.
\newblock Better mixing via deep representations.
\newblock In \emph{International conference on machine learning}, pages
  552--560. PMLR, 2013.

\bibitem[Berthelot et~al.(2018)Berthelot, Raffel, Roy, and
  Goodfellow]{berthelot2018understanding}
David Berthelot, Colin Raffel, Aurko Roy, and Ian Goodfellow.
\newblock Understanding and improving interpolation in autoencoders via an
  adversarial regularizer.
\newblock \emph{arXiv preprint arXiv:1807.07543}, 2018.

\bibitem[Chen et~al.(2018)Chen, Li, Grosse, and Duvenaud]{chen2018isolating}
Ricky T.~Q. Chen, Xuechen Li, Roger Grosse, and David Duvenaud.
\newblock Isolating sources of disentanglement in variational autoencoders.
\newblock In \emph{Advances in Neural Information Processing Systems}, 2018.

\bibitem[Chen et~al.(2016)Chen, Duan, Houthooft, Schulman, Sutskever, and
  Abbeel]{chen2016infogan}
Xi~Chen, Yan Duan, Rein Houthooft, John Schulman, Ilya Sutskever, and Pieter
  Abbeel.
\newblock Infogan: Interpretable representation learning by information
  maximizing generative adversarial nets.
\newblock \emph{arXiv preprint arXiv:1606.03657}, 2016.

\bibitem[Engel et~al.(2017)Engel, Resnick, Roberts, Dieleman, Norouzi, Eck, and
  Simonyan]{engel2017neural}
Jesse Engel, Cinjon Resnick, Adam Roberts, Sander Dieleman, Mohammad Norouzi,
  Douglas Eck, and Karen Simonyan.
\newblock Neural audio synthesis of musical notes with wavenet autoencoders.
\newblock In \emph{International Conference on Machine Learning}, pages
  1068--1077. PMLR, 2017.

\bibitem[Eom and Ham(2019)]{eom2019learning}
Chanho Eom and Bumsub Ham.
\newblock Learning disentangled representation for robust person
  re-identification.
\newblock \emph{arXiv preprint arXiv:1910.12003}, 2019.

\bibitem[Ge et~al.(2020{\natexlab{a}})Ge, Abu-El-Haija, Xin, and
  Itti]{ge2020zero}
Yunhao Ge, Sami Abu-El-Haija, Gan Xin, and Laurent Itti.
\newblock Zero-shot synthesis with group-supervised learning.
\newblock \emph{arXiv preprint arXiv:2009.06586}, 2020{\natexlab{a}}.

\bibitem[Ge et~al.(2020{\natexlab{b}})Ge, Zhao, and Itti]{ge2020pose}
Yunhao Ge, Jiaping Zhao, and Laurent Itti.
\newblock Pose augmentation: Class-agnostic object pose transformation for
  object recognition.
\newblock In \emph{European Conference on Computer Vision}, pages 138--155.
  Springer, 2020{\natexlab{b}}.

\bibitem[Higgins et~al.(2017)Higgins, Matthey, Pal, Burgess, Glorot, Botvinick,
  Mohamed, and Lerchner]{Higgins2017betaVAELB}
I.~Higgins, Lo{\"i}c Matthey, A.~Pal, C.~Burgess, Xavier Glorot, M.~Botvinick,
  S.~Mohamed, and Alexander Lerchner.
\newblock beta-vae: Learning basic visual concepts with a constrained
  variational framework.
\newblock In \emph{ICLR}, 2017.

\bibitem[Higgins et~al.(2016)Higgins, Matthey, Pal, Burgess, Glorot, Botvinick,
  Mohamed, and Lerchner]{higgins2016beta}
Irina Higgins, Loic Matthey, Arka Pal, Christopher Burgess, Xavier Glorot,
  Matthew Botvinick, Shakir Mohamed, and Alexander Lerchner.
\newblock beta-vae: Learning basic visual concepts with a constrained
  variational framework.
\newblock 2016.

\bibitem[Karras et~al.(2019)Karras, Laine, and Aila]{karras2019style}
Tero Karras, Samuli Laine, and Timo Aila.
\newblock A style-based generator architecture for generative adversarial
  networks.
\newblock In \emph{Proceedings of the IEEE/CVF Conference on Computer Vision
  and Pattern Recognition}, pages 4401--4410, 2019.

\bibitem[Kingma and Welling(2014)]{kingma2014autoencoding}
Diederik~P Kingma and Max Welling.
\newblock Auto-encoding variational bayes, 2014.

\bibitem[Lee et~al.(2018)Lee, Tseng, Huang, Singh, and Yang]{lee2018diverse}
Hsin-Ying Lee, Hung-Yu Tseng, Jia-Bin Huang, Maneesh Singh, and Ming-Hsuan
  Yang.
\newblock Diverse image-to-image translation via disentangled representations.
\newblock In \emph{Proceedings of the European conference on computer vision
  (ECCV)}, pages 35--51, 2018.

\bibitem[Likas et~al.(2003)Likas, Vlassis, and Verbeek]{likas2003global}
Aristidis Likas, Nikos Vlassis, and Jakob~J Verbeek.
\newblock The global k-means clustering algorithm.
\newblock \emph{Pattern recognition}, 36\penalty0 (2):\penalty0 451--461, 2003.

\bibitem[Liu et~al.(2015)Liu, Luo, Wang, and Tang]{liu2015deep}
Ziwei Liu, Ping Luo, Xiaogang Wang, and Xiaoou Tang.
\newblock Deep learning face attributes in the wild.
\newblock In \emph{Proceedings of the IEEE international conference on computer
  vision}, pages 3730--3738, 2015.

\bibitem[Mehrabi et~al.(2019)Mehrabi, Morstatter, Saxena, Lerman, and
  Galstyan]{mehrabi2019survey}
Ninareh Mehrabi, Fred Morstatter, Nripsuta Saxena, Kristina Lerman, and Aram
  Galstyan.
\newblock A survey on bias and fairness in machine learning, 2019.

\bibitem[Ronneberger et~al.(2015)Ronneberger, Fischer, and
  Brox]{ronneberger2015u}
Olaf Ronneberger, Philipp Fischer, and Thomas Brox.
\newblock U-net: Convolutional networks for biomedical image segmentation.
\newblock In \emph{International Conference on Medical image computing and
  computer-assisted intervention}, pages 234--241. Springer, 2015.

\bibitem[Sainburg et~al.(2018)Sainburg, Thielk, Theilman, Migliori, and
  Gentner]{sainburg2018generative}
Tim Sainburg, Marvin Thielk, Brad Theilman, Benjamin Migliori, and Timothy
  Gentner.
\newblock Generative adversarial interpolative autoencoding: adversarial
  training on latent space interpolations encourage convex latent
  distributions.
\newblock \emph{arXiv preprint arXiv:1807.06650}, 2018.

\bibitem[Selvaraju et~al.(2017)Selvaraju, Cogswell, Das, Vedantam, Parikh, and
  Batra]{selvaraju2017grad}
Ramprasaath~R Selvaraju, Michael Cogswell, Abhishek Das, Ramakrishna Vedantam,
  Devi Parikh, and Dhruv Batra.
\newblock Grad-cam: Visual explanations from deep networks via gradient-based
  localization.
\newblock In \emph{Proceedings of the IEEE international conference on computer
  vision}, pages 618--626, 2017.

\bibitem[Shen et~al.(2020)Shen, Yang, Tang, and Zhou]{shen2020interfacegan}
Yujun Shen, Ceyuan Yang, Xiaoou Tang, and Bolei Zhou.
\newblock Interfacegan: Interpreting the disentangled face representation
  learned by gans.
\newblock \emph{arXiv preprint arXiv:2005.09635}, 2020.

\bibitem[Tran et~al.(2017)Tran, Yin, and Liu]{tran2017disentangled}
Luan Tran, Xi~Yin, and Xiaoming Liu.
\newblock Disentangled representation learning gan for pose-invariant face
  recognition.
\newblock In \emph{Proceedings of the IEEE conference on computer vision and
  pattern recognition}, pages 1415--1424, 2017.

\bibitem[Verma et~al.(2018)Verma, Lamb, Beckham, Courville, Mitliagkis, and
  Bengio]{verma2018manifold}
Vikas Verma, Alex Lamb, Christopher Beckham, Aaron Courville, Ioannis
  Mitliagkis, and Yoshua Bengio.
\newblock Manifold mixup: Encouraging meaningful on-manifold interpolation as a
  regularizer.
\newblock \emph{arXiv preprint arXiv:1806.05236}, 7, 2018.

\bibitem[Xiao et~al.(2017)Xiao, Hong, and Ma]{xiao2017dna}
Taihong Xiao, Jiapeng Hong, and Jinwen Ma.
\newblock Dna-gan: Learning disentangled representations from multi-attribute
  images.
\newblock \emph{arXiv preprint arXiv:1711.05415}, 2017.

\bibitem[Xiao et~al.(2018)Xiao, Hong, and Ma]{xiao2018elegant}
Taihong Xiao, Jiapeng Hong, and Jinwen Ma.
\newblock Elegant: Exchanging latent encodings with gan for transferring
  multiple face attributes.
\newblock In \emph{Proceedings of the European Conference on Computer Vision
  (ECCV)}, pages 172--187, September 2018.

\bibitem[Zheng et~al.(2019)Zheng, Yang, Yu, Zheng, Yang, and
  Kautz]{zheng2019joint}
Zhedong Zheng, Xiaodong Yang, Zhiding Yu, Liang Zheng, Yi~Yang, and Jan Kautz.
\newblock Joint discriminative and generative learning for person
  re-identification.
\newblock In \emph{Proceedings of the IEEE/CVF Conference on Computer Vision
  and Pattern Recognition}, pages 2138--2147, 2019.

\end{thebibliography}

\clearpage
\appendix
\section*{Appendix}

\section{Network Architecture and Training Details}
\subsection{ELEGANT \cite{xiao2018elegant} + CIR}
\noindent{\bf {Network Structure}}

For our ELEGANT + Controllable Interpolation Regularization (CIR), we use the same network architecture as the original ELEGANT paper \cite{xiao2018elegant}. We use an autoencoder-structure generator $G$ with an encoder $E$ and a decoder $D$. The $E$ and $D$ structures are symmetrical with an architecture consisting of five convolutional layers. As for the discriminator $D$, it adopts multi-scale discriminators $D1$ and $D2$. Both $D1$ and $D2$ use a CNN architecture with four convolutional layers followed by a fully-connected layer. The difference between them is that $D1$ has a larger fully-connected layer while the one of $D2$'s is small.

\noindent{\bf {Training Details}}

We train the ELEGANT and ELEGANT + CIR models on \textit{CelebA} \cite{liu2015deep}.  The size of input images is $256 \times 256$. Both generator and discriminator use Adam with $\beta_{1}$=0.5 and $\beta_{2}$=0.999, batch size 16, learning rate 0.0002 at first and multiply 0.97 every 3000 epochs.

Hyperparameters in the loss function: For reconstruction loss and Adversarial loss, we use $\lambda_{reconstruction}=5$, $\lambda_{adversarial}=1$ unmodified. For Controllable Interpolation Regularization loss, we set $\lambda_{CIR}=1 \times 10^{7}$ to make the regularization loss has a similar scale as other loss terms and balance the training.

Disentangle details: The encoder of generator $G$ maps an image into a latent code with shape $(512 \times 8 \times 8)$, and ELEGANT will dynamically allocate these spaces to store information of the interesting attributes. For instance, suppose the attributes we want to disentangle are eyeglasses and mustache. Then the input will be [eyeglasses, mustache], and the first half of latent space will store the information of eyeglasses. In other words, we disentangle the latent space along the first dimension and both eyeglasses and mustache get $(256 \times 8 \times 8)$ latent space.

\subsection{I2I-Dis \cite{lee2018diverse} + CIR}

\noindent{\bf {Network Structure}}

We use the same network architecture as the original I2I-Dis paper \cite{lee2018diverse}. For all the experiments in this section, we use images from \textit{cat2dog} dataset with size $216 \times 216$. There are four modules in I2I-Dis: shared content encoder $E^c$, domain-specific attribute encoder $E^a$, generator $G$, discriminator $D$. For the shared content encoder $E^c$, we use an architecture consisting of three convolutional layers followed by four residual blocks. For the domain-specific attribute encoder $E^a$, we use a CNN architecture with four convolutional layers followed by fully-connected layers. For the generator $G$, we use an architecture consisting of four residual blocks followed by three fractionally stridden convolutional layers. For the discriminator $D$, we use an architecture consisting of four convolutional layers followed by fully-connected layers. Our disentangled latent code consists of two-part: shared content attribute latent code $z_c$ with shape $256\times 54\times 54$ and domain-specific attribute latent code $z_a$ with shape $8 \times 1$.

\noindent{\bf {Training Details}}

The training of I2I-Dis and I2I-Dis + CIR use Adam optimizer with batch size of 1, learning rate of 0.0001, and exponential decay rates $\beta_1=0.5,\beta_2=0.999$. 

Hyper-parameters in loss function: For reconstruction loss, we use $\lambda_1^{rec}=10,\lambda_{cc}=10$. For adversarial loss, we use $\lambda_{adv}^{content}=1,\lambda_{adv}^{domain}=1$. For latent regression loss, we use $\lambda_1^{latent}=10$. For KL divergence loss, we use $\lambda_{KL}=0.01$. For our controllable interpolation regularization loss, we use $\lambda_{CIR}=10$.

\subsection{GZS-Net \cite{ge2020zero} + \textbf{CIR}}
\noindent{\bf {Network Structure}}

We use the same network architecture and the same dataset (\textit{Fonts} \cite{ge2020zero}) as the original GZS-Net paper \cite{ge2020zero}. The input images are of size $128 \times 128$. There are two modules in GZS-Net: an encoder $E$ and a decoder $D$. The \textit{Fonts} dataset have 5 attributes: content, size, font color, background color and font. Each attribute takes 20 dimensions in the latent space and thereby sums up to a 100-dimensional vector. The encoder $E$ is composed of two convolutional layers with stride 2, followed by three residual blocks. Then it comes with a convolutional layer with stride 2, followed by a flatten layer that reshapes the response map to a vector. Finally, two fully-connected layers output 100-dimensional vectors as the latent feature. The decoder $D$ mirrors the encoder, composed of two fully-connected layers, followed by a cuboid-reshaping layer. The next is a deconvolutional layer with stride 2, followed by three residual blocks. And finally, two deconvolutional layers with stride 2 produce a synthesized image.

\noindent{\bf{Training Details}}

We train GZS-Net and GZS-Net + CIR on \textit{Fonts} \cite{ge2020zero} dataset. We use Adam optimizer with batch size of 8, learning rate of 0.0001, and exponential decay rates $\beta_1=0.9,\beta_2=0.999$.

Hyper-parameters in loss function: For reconstruction loss, we use $\lambda_1^{rec}=1,\lambda_{combine}=1$. For our controllable interpolation regularization loss, we use $\lambda_{CIR}=0.0001$ at an early stage and $\lambda_{CIR}=0.01$ after 100000 epochs to balance the training.

\section{More Qualitative Results}

\begin{figure*}[htbp]
  \centering
  \includegraphics[width=\linewidth]{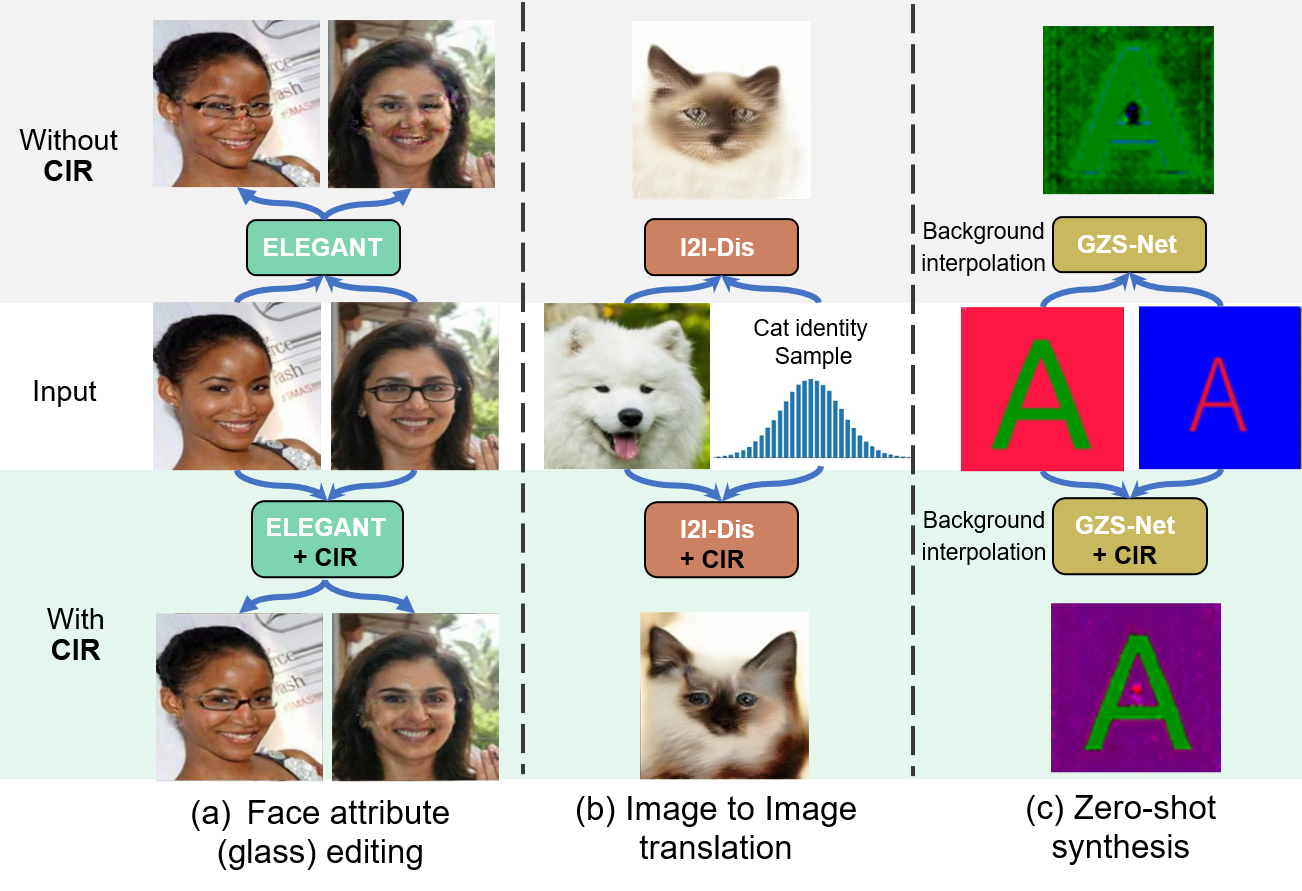}
  \caption{
  Our proposed approach CIR improves the result quality of 3 tasks by encouraging both disentanglement and convexity in the latent space: (a) Face attribute editing with ELEGANT (add/remove glasses on face); CIR is better able to transfer glasses with less disturbance on other face parts. (b) Image to image translation transfer from a dog image to a cat image with same pose (content); CIR better matches the desired pose with fewer artifacts. (c) Zero-shot synthesis with GZS-Net to synthesize an image with a new background by interpolating in the corresponding latent space; CIR better interpolates the background only without changing letter size, color or font style.}
  \label{fig:1-large}
\end{figure*}

Fig.~\ref{fig:1-large} shows a larger version of main paper Fig.~1 to show more details.

\begin{figure*}
  \centering
  \includegraphics[width=\linewidth]{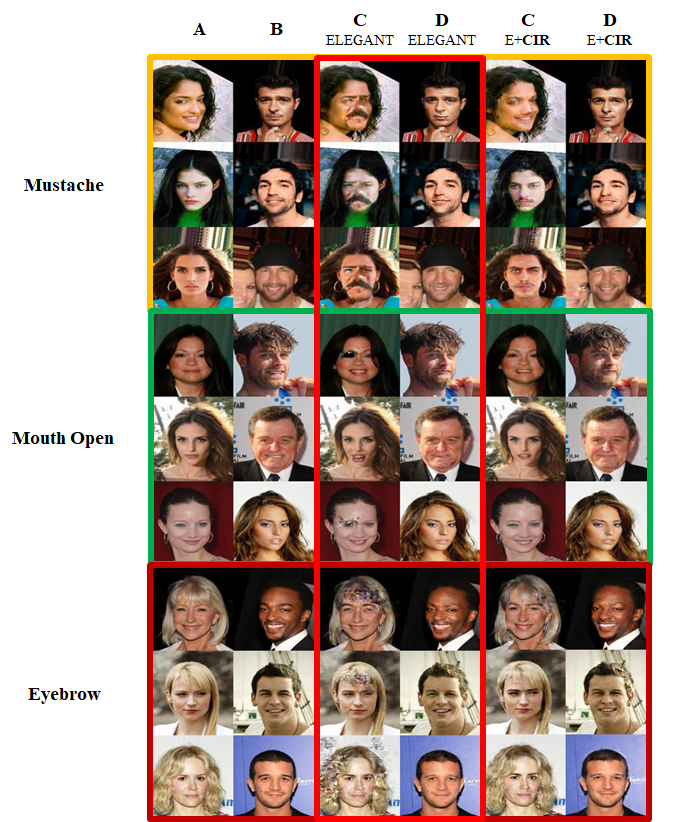}
  \caption{
  More examples of ELEGANT+\textbf{CIR} (E+\textbf{CIR}) performance of task 1 for two images face attribute transfer}
  \label{fig:supp-C-3}
  \vspace{-10pt}
\end{figure*}

\begin{figure*}
\begin{center}
\includegraphics[width=\linewidth]{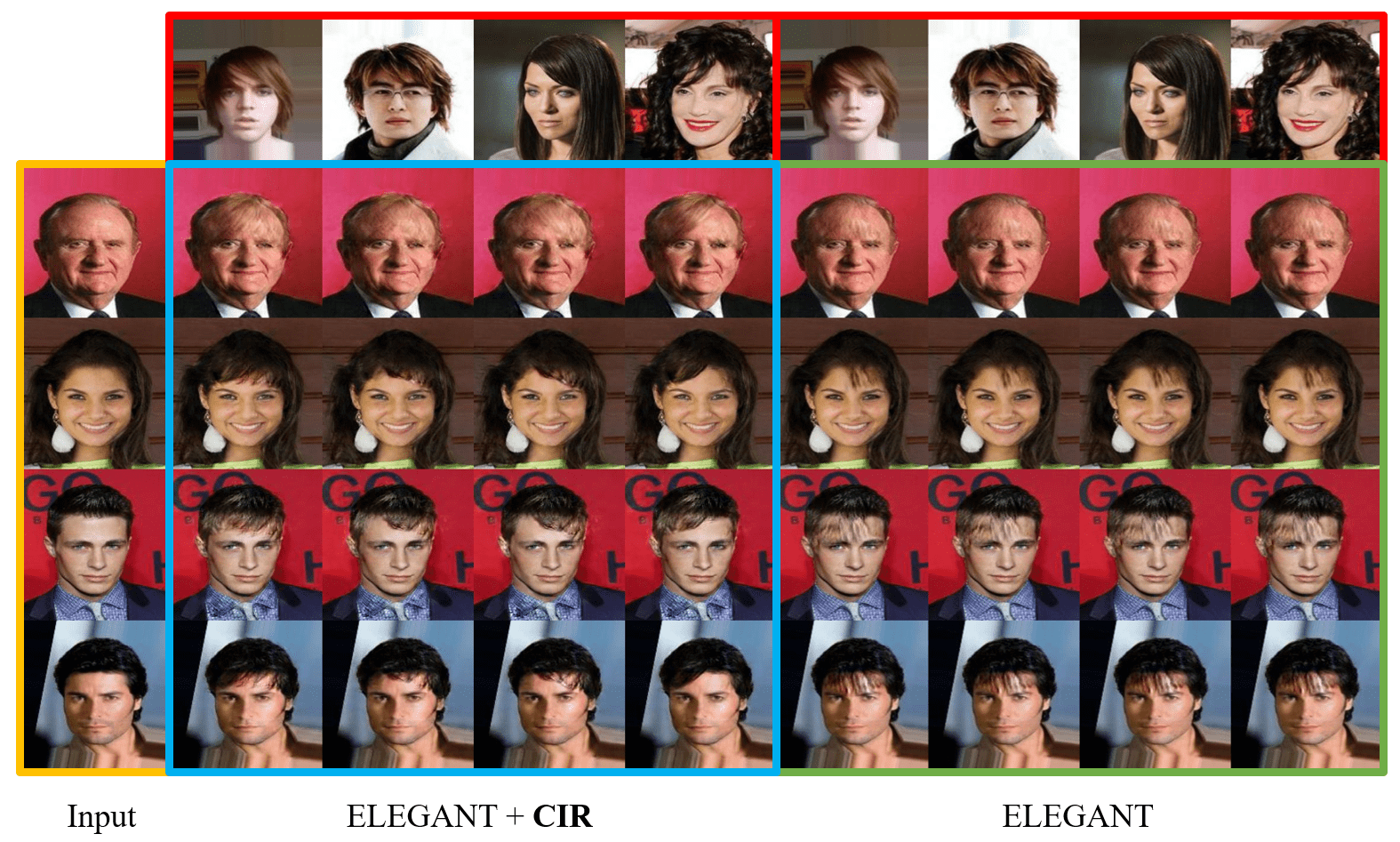}
\end{center}
\setlength{\abovecaptionskip}{2pt}
  \caption{ELEGANT + \textbf{CIR} Performance of task 2 for face image generation by exemplars}
\label{fig:supp-B-3}
\end{figure*}

\subsection{ELEGANT \cite{xiao2018elegant} + CIR}

Fig.~\ref{fig:supp-C-3} shows more results of the task 1 performance on two images face attribute transfer, which is similar to the main paper Fig. 3. We offer three rows for each attribute, including a new attribute (Mouth-Open vs. Mouth-Close). 

Fig.~\ref{fig:supp-B-3} shows  more results of the task 2 performance on face image generation by exemplars, which is similar to the main paper Fig. 4 but with bangs as our disentangle attribute. The results show that CIR can help to overcome the mode collapse problem in ELEGANT.

\subsection{I2I-Dis \cite{lee2018diverse} + CIR}
\begin{figure*}[htbp]
\begin{center}
\includegraphics[width=\linewidth]{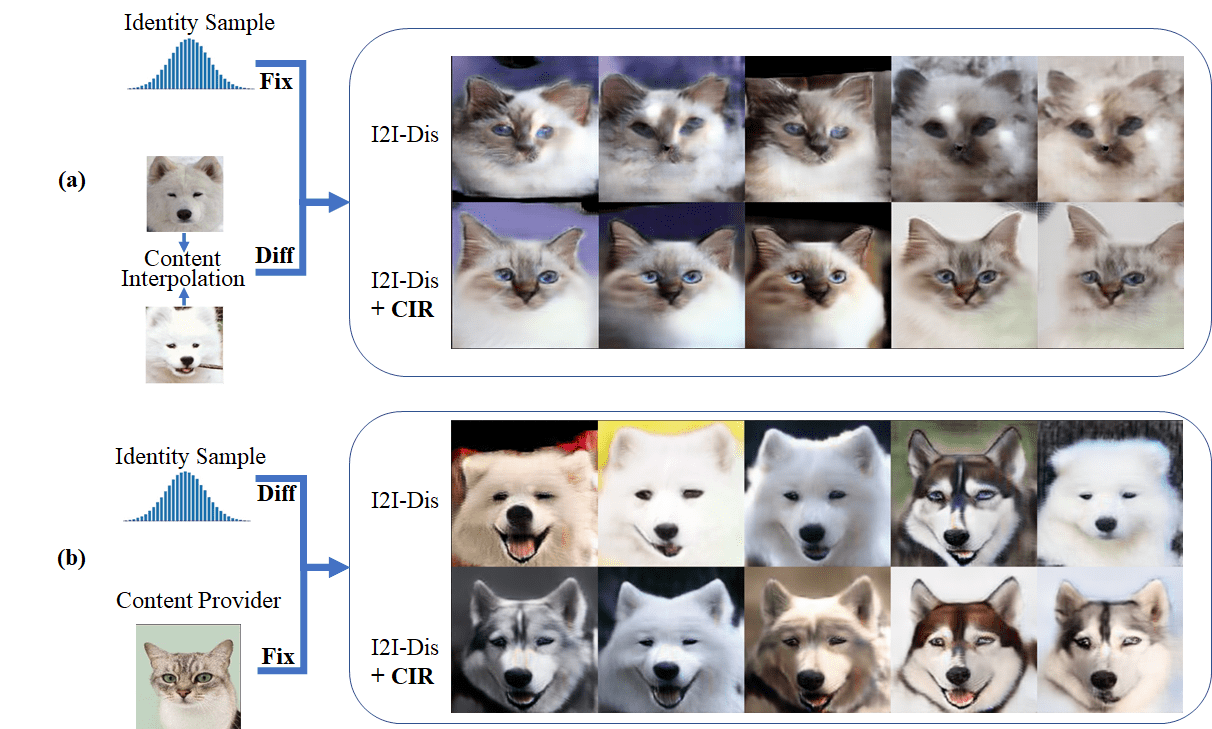}
\end{center}
\setlength{\abovecaptionskip}{2pt}
  \caption{I2I-Dis + \textbf{CIR} performance of diverse image-to-image translation}
\label{fig:fig-11}
\end{figure*}

Fig.~\ref{fig:fig-11} shows more results of the image-to-image translation, which is similar to the main paper Fig. 5. (a) We generate cat images given fixed identity (domain) attribute latent code and change the 'content' attribute latent code by interpolation. (b) We generate dog images given fixed content attribute latent code and change the 'identity' attribute latent code by sampling.

\vspace{-2pt}
\subsection{GZS-Net \cite{ge2020zero} + \textbf{CIR}}
\begin{figure*}[htbp]
  \centering
  \includegraphics[width=\linewidth]{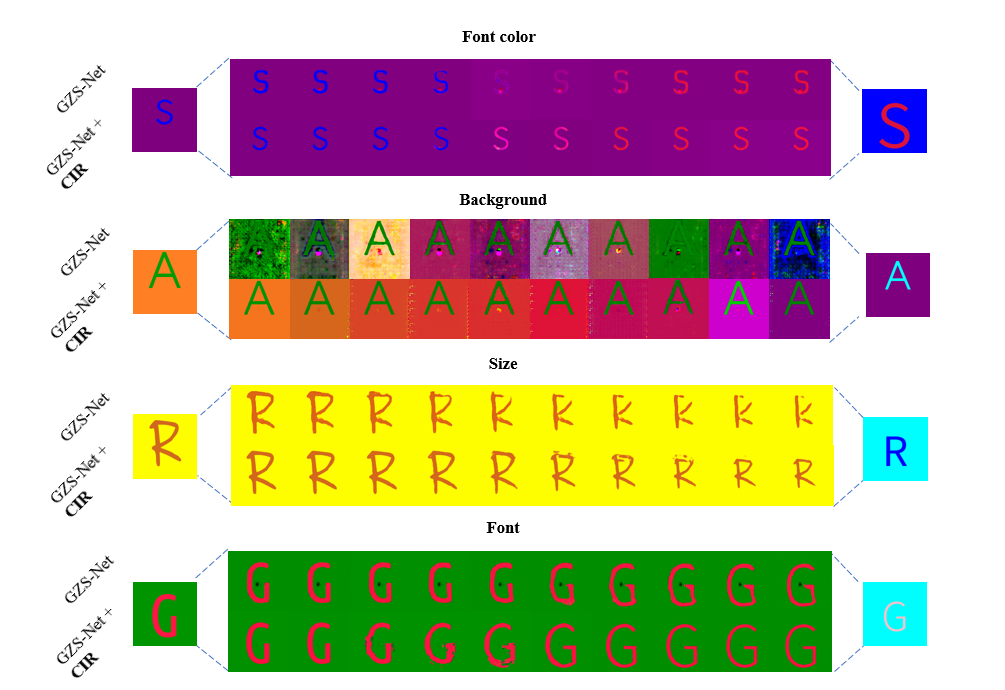}
  \caption{
  More results of GZS-Net + \textbf{CIR} performance of interpolation-based attribute controllable synthesis}
  \label{fig:GSL_more_results}
\end{figure*}

Fig.~\ref{fig:GSL_more_results} shows more results of the interpolation-based controllable synthesis performance on font color, background color, size, and font attributes.



\section{Quantitative Experiments Details}

\subsection{Disentanglement Evaluation by Correlation Coefficient.}

We use Spearman's Rank Correlation for latent space correlation computation. It is computed as:
\begin{equation}\label{Eq.a1}
	r_s=\frac{cov(rg_{X}, rg_{Y})}{\sigma_{rg_{X}}\sigma_{rg_{Y}}}
\end{equation}
Here $rg_{X}$ and $rg_{Y}$ means the rank variables of $X$ and $Y$. $cov$ is the covariance function. $\sigma$ denotes the standard variation.

For ELEGANT + CIR that disentangles eyeglasses and mustache, we collect 10,000 images from \textit{CelebA} \cite{liu2015deep} and obtain the same number of $(512 \times 8 \times 8)$ latent matrices from encoder. Then we average the vectors along the $2^{nd}$ and $3^{rd}$ dimensions and produce squeezed matrices of size 512. This preprocessing step is following the interpolation strategy, which helps to display the intra correlation more clearly.

For GZS-Net + CIR, 10,000 images are fetched from \textit{Fonts} and corresponding latent vectors with size 100 are computed. No preprocessing is applied.

All the latent matrices (or vectors) are normalized before putting into Spearman's Rank Correlation calculation. The normalization is calculated as:
\begin{equation}\label{Eq.a2}
	norm(v_i) = (v_i - \overline{v_i})~/~\sigma_{v_{i}},~ \forall~ i \in \{1, 2, ..., |v|\}
\end{equation}
$v_i$ is the value of each dimension $i$ in $v$. $\overline{v_i}$ is the average of $v_i$ and $\sigma_{v_{i}}$ is the standard variance.

\begin{figure*}[htbp]
  \centering
  \includegraphics[width=\linewidth]{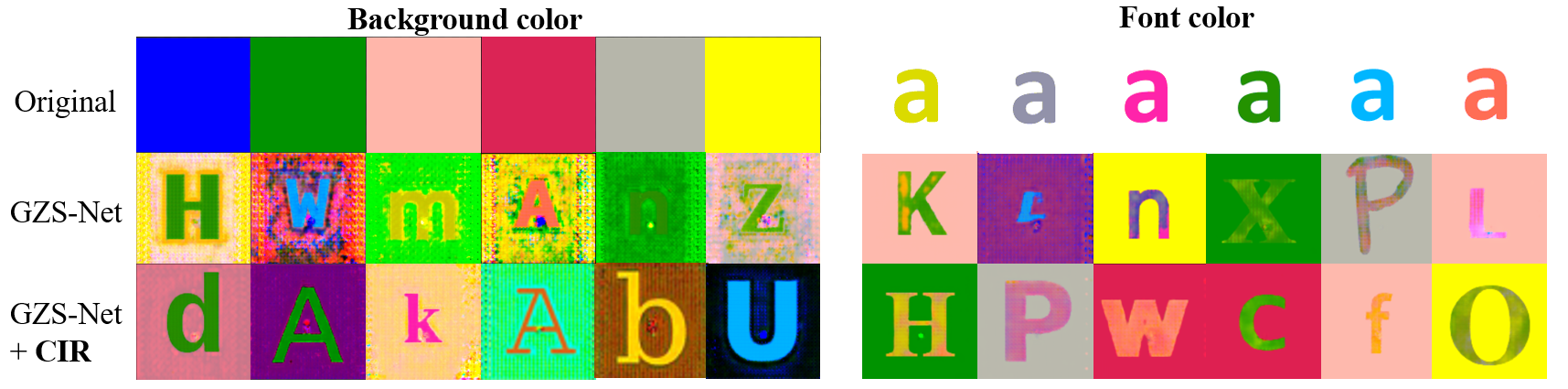}
  \caption{
  Controllable mining novel background and font color by interpolation in latent space. }
  \label{fig:mining}
\end{figure*}

\begin{figure*}[htbp]
  \centering
  \includegraphics[width=0.7\linewidth]{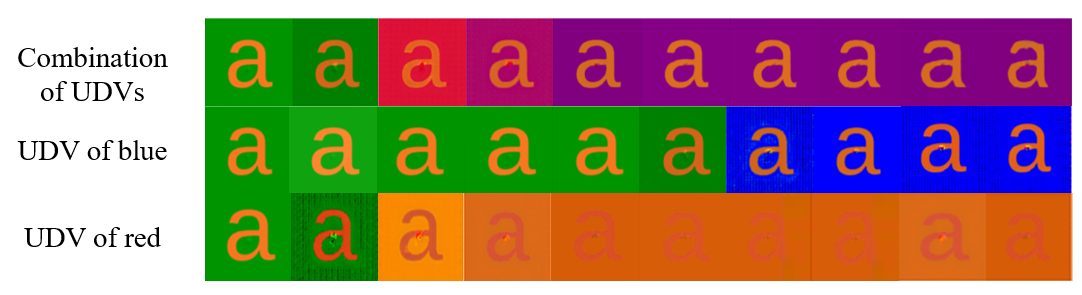}
  \caption{
  Mining new attribute values with UDV}
  \label{fig:supp-C-2}
\end{figure*}


\vspace{-2pt}
\section{More Downstream Tasks and Applications}
\vspace{-2pt}
\label{exp:6}
We conduct more experiments to demonstrate 3 potential applications with the more convex and robust disentangled latent space by CIR.

\noindent{\bf Data Augmentation.}
We design a letter image classification experiment with \textit{Fonts} \cite{ge2020zero} to evaluate how interpolation-based controllable synthesis ability, empowered by CIR, as a data augmentation method, improves the downstream classification task. 
We tailored three datasets from \textit{Fonts}, each of them has ten letters as labels. The large training set ($D_{L}$) and testing set ($D_{test}$) have the same number of images with the same attribute values. We take a subset of $D_{L}$ to form a small training set $D_{S}$ with fewer attribute values.
For data augmentation, we first train the GZS-Net and GZS-Net + CIR on $D_{S}$, and then we use the trained models to generate 1000 new images by interpolation-based attribute controllable synthesis. We combine the synthesized images with $D_{S}$ and form two augmented training sets $D_{S+G}$ (GZS-Net) and $D_{S+G+C}$ (GZS-Net + CIR), respectively.  All test accuracy shown in (Table~\ref{table-4}), which shows an improved data augmentation performance on downstream tasks with the help of CIR. (more details in Supplementary)

\begin{table}[t]
\scriptsize {}
\caption{Controllable augmentation performance (the $\star$ means that synthesized images with new attributes are added into training set)}
\label{table-4}
\begin{tabular}{l|c|c|c|c|c}
\hline
 \diagbox[width=9.6em]{Attribute}{Dataset}
 & $D_{L}$ & $D_{S}$ & $D_{S+G}$ & $D_{S+G + C}$
 & $D_{test}$ \\

\hline
 Size & 3 & 2 & 2$\star$ & 2$\star$ & 3 \\
 Font Color & 6 & 3 & 3$\star$ & 3$\star$ & 6 \\
 Back Color & 3 & 3 & 3$\star$ & 3$\star$ & 3 \\
 Fonts & 10 & 3 & 3$\star$ & 3$\star$ & 10 \\
 Dataset Size & 5400 & 540 & 540+1000 & 540+1000 & 5400 \\
 Train Accuracy & 98\% & 99\% & 99\% & 99\% & N/A \\
 Test Accuracy & 94\% & 71\% & 74\% & \textbf{76\%} & N/A \\
\hline

\end{tabular}
\vspace{-10pt}
\end{table}

\noindent{\bf Bias Elimination for Fairness.}
Dataset bias may influence the model performance significantly. \cite{mehrabi2019survey} listed lots of bias resources and proved that eliminating bias is significant. A more convex and disentangled representation with CIR could be a solution to the bias problem by first disentangle the bias attribute and then remove them in the final decision.
We use the \textit{Fonts} dataset to simulate the bias problem.
We tailored three datasets, a biased training dataset $\mathcal{D^\textrm{B}}$, two unbiased dataset: $\mathcal{D^\textrm{UB}}$ for training and $\mathcal{D^\textrm{T}}$ for test. In $\mathcal{D^\textrm{B}}$, we entangle the two attributes, letter and background color, as dataset bias. $\mathcal{D^\textrm{B}}$ consists of three-part: G1, G2, and G3, where each letter has 1, 3, and 6 background colors, respectively. (more details in Supplementary) 
Then, we use $\mathcal{D^\textrm{B}}$ and $\mathcal{D^\textrm{UB}}$ to train letter classifier with resnet-18 respectively and test on $\mathcal{D^\textrm{T}}$ as the control groups. 
As is shown in Table.~\ref{table-6}, the classifier trained on $\mathcal{D^\textrm{B}}$, only gets 81\% test accuracy while classifier trained on $\mathcal{D^\textrm{UB}}$ obtains 99\% test accuracy. 
As shown in Fig.~\ref{fig:9}, Grad-Cam's \cite{selvaraju2017grad} results proved that the classifier would regard background color as essential information if it entangled with letters. 
We use the more convex and disentangled representation of CIR to solve the entangled bias in $\mathcal{D^\textrm{B}}$. We first train a GZS-Net + CIR use $\mathcal{D^\textrm{B}}$. Then we train a letter classifier on the latent representation instead of image space, where we explicitly drop the background color-related dimensions (bias attribute) and use the rest of the latent code as input.  After training, the accuracy rises to 98\%. Hence, we eliminate the dataset bias with the help of robust disentangled latent by CIR.




\begin{table}[t]
\centering
\scriptsize
\setlength{\abovecaptionskip}{3pt}
\caption{Bias elimination experiment results}
\label{table-6}
\begin{tabular}{c|c|c|c}
\hline
\diagbox{Dataset}{Model} & \makecell {resnet18  $\mathcal{D^\textrm{B}}$} & \makecell {resnet18  $\mathcal{D^\textrm{UB}}$} & \makecell {GZS-Net + CIR  $\mathcal{D^\textrm{B}}$} \\
\hline
Test(Letters in G1) & 52.73\% & 99.17\% & 96.77\%\\
\hline
Test(Letters in G2) & 82.63\% & 98.67\% & 98.97\%\\
\hline
Test(Letters in G3) & 99.13\% & 98.30\% & 98.46\%\\
\hline
Train & 99.44\% & 98.82\% & 99.98\%\\
\hline
Test & \textbf{81.32\%} & 98.63\% & \textbf{98.11\%}\\
\hline

\end{tabular}
\end{table}

\begin{figure}[t]
\vspace{-5pt}
  \centering
 \includegraphics[width=0.7\linewidth]{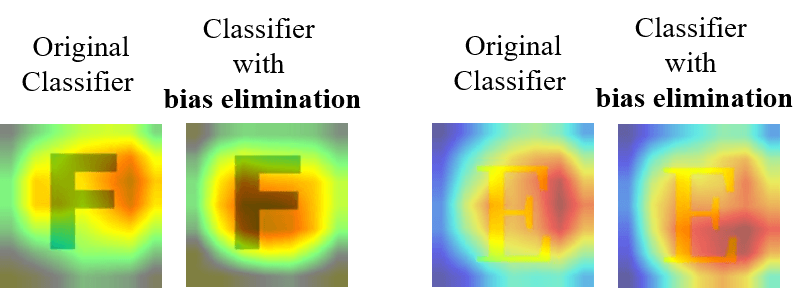}
  \caption{
  The influence of bias shown by Grad-Cam}
  \label{fig:9}
  \vspace{-15pt}
\end{figure}

\begin{table}[h]
\centering
\scriptsize
\caption{Bias elimination dataset setting}
\label{table-supp-C-1}
\begin{tabular}{cc|c|c}
\hline
 \multicolumn{2}{c|}{Dataset}  & \makecell[c]{Number of\\ letters}  & \makecell[c]{Number of\\ colors}  \\
\hline
\multirow{3}{*}{$\mathcal{D^\textrm{B}}$} & G1 & 15 & 1 \\
& G2 & 15 & 3 \\
& G3 & 22 & 6 \\
\hline
\multicolumn{2}{c|}{$\mathcal{D^\textrm{UB}}$} & 52 & 6 \\
\hline
\multicolumn{2}{c|}{$\mathcal{D^\textrm{T}}$} & 52 & 6\\
\hline
\end{tabular}
\end{table}


\noindent{\bf Mining New Attribute Value.} Fig.\ref{fig:mining} shows our results of mining new attribute value. To find a good exploration direction and mine new attribute values, we explore the distribution of each attribute value in the corresponding attribute-convex latent space (e.g., the distribution of different background colors in a convex background color latent space: $\mathcal{A}_{back} = \{\textrm{blue}, \textrm{red}, \textrm{green}, \textrm{yellow},\dots \}$).

\begin{figure}[htbp]
  \centering
  \includegraphics[width=\linewidth]{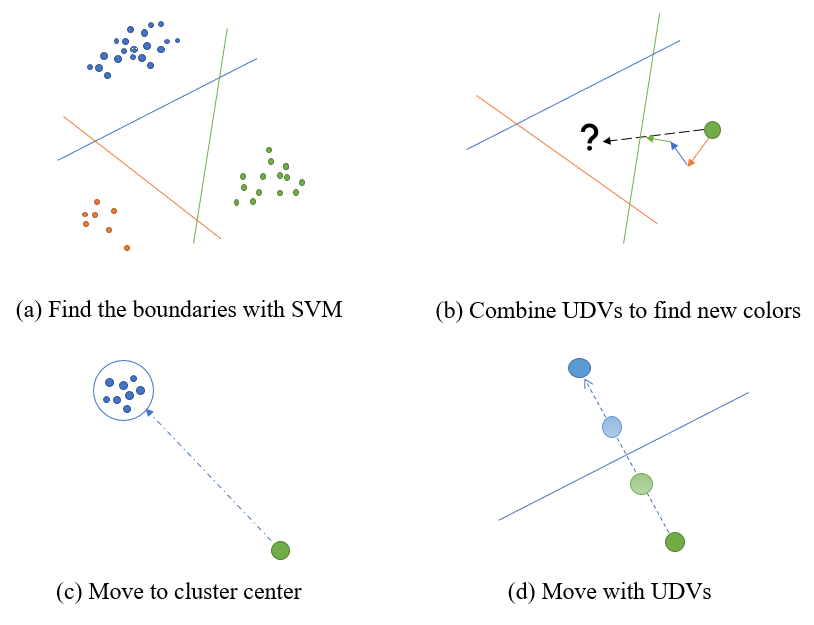}
  \caption{
  Towards controllable exploration direction}
  \label{fig:supp-C-1}
\end{figure}

Two common kinds of distribution are considered: \\
1) \textbf{Gaussian}. For those attributes (object color) whose attribute value (blue color) has slight intra-class variance (all blues look similar), their distribution can be seen as a Gaussian distribution. We can use K-means \cite{likas2003global} to find the center of each object color and guide the interpolation and synthesis. \\
2) \textbf{Non-Gaussian}. We treat each attribute value as a binary semantic label (e.g., wear glasses or not wear glasses ). We assume a hyperplane in the latent space serving as the separation boundary \cite{shen2020interfacegan}, and the distance from a sample to this hyperplane is in direct proportion to its semantic score. We can train an SVM to find this boundary and use the vector orthogonal to the border and the positive side to represent a Unit Direction Vector (UDV). We can then use the UDVs or a combination to achieve precise attribute synthesis and find new attribute values.
As shown in Fig.~\ref{fig:supp-C-1} (a), we can find the boundaries and UDVs by SVM for each attribute value. To solve the precision problem in attribute synthesis, Fig.~\ref{fig:supp-C-1} (c) shows moving towards the z value of the cluster center directly for Gaussian; Fig.~\ref{fig:supp-C-1} (d) shows moving from the start point, across the boundary, to the target attribute value, by adding the UDV of the target attribute for non-Gaussian. 
Fig.~\ref{fig:supp-C-1} (b) shows that we can combine the UDVs to discover new attribute values.

Here we explore the distribution of disentangled representation and mining the relationship between movement in high dimension $x$ space and low dimension $z$ space to answer the question: Which direction of movement can help us to find new attributes?

For each background color, we train a binary color classifier to label interpolated points in the $z$ space and assign a color score for each of them. Then we use SVM to find the boundary and obtain UDV for this attribute value. Since the UDV is the most effective direction to change the semantic score of samples, if we move $z$ value of the given image towards UDV, its related semantic score would increase fast. To explore more new attributes, the combination of UDVs may be a good choice. For instance, if the given picture is green, the new colors may fall in the path from green to blue and the path from green to red. Thus, it is reasonable to set our move direction as $v =v_{blue}+v_{red}-v_{green}$ ($v$ represents UDV). 
The 1$^{st}$ row of Fig.~\ref{fig:supp-C-2} shows the results of changing $z$ value with the combine vector $v_{blue}+v_{red}-v_{green}$. On the contrast, the 2$^{nd}$ row only use $v_{blue}$ and the 3$^{rd}$ row only use $v_{red}$. We can find that both the 2$^{nd}$ and the 3$^{rd}$ row only find one color while the 1$^{st}$ row finds more.






~\newpage



\providecommand{\upGamma}{\Gamma}
\providecommand{\uppi}{\pi}

\end{document}